\documentclass[10pt,twocolumn,letterpaper]{article}

\usepackage{cvpr}
\usepackage{times}
\usepackage{epsfig}
\usepackage{graphicx}
\usepackage{amsmath}
\usepackage{amssymb}
\usepackage{adjustbox}
\usepackage{subcaption}
\usepackage{bm}
\usepackage{caption}
\usepackage{booktabs}
\usepackage{array}
\usepackage{tasks}

\usepackage[pagebackref=true,breaklinks=true,letterpaper=true,colorlinks,bookmarks=false]{hyperref}

\cvprfinalcopy % *** Uncomment this line for the final submission

\def\cvprPaperID{1024} % *** Enter the CVPR Paper ID here

\newcommand{\hamedrem}[1]{\textcolor{red}{}}

\ifcvprfinal\fi
\begin{document}

%%%%%%%%% TITLE
\title{Learning to Transfer with Incompatibility}
\title{A Novel Distillation Method for Self-Supervised Learning}
\title{Transferring Knowledge for Self-Supervised Learning}
\title{Boosting Self-Supervised Learning via Knowledge Transfer}
%\title{Boosting Self-Supervised Learning via Knowledge Distillation}

\author{Mehdi Noroozi$^1\quad$ Ananth Vinjimoor$^2 \quad$ Paolo Favaro$^1\quad$ Hamed Pirsiavash$^2$\\
\hspace{0em} University of Bern$^1\quad   $ University of Maryland, Baltimore County$^2$\\
{\hspace{-0em} $\texttt{\small \{noroozi,favaro\}@inf.unibe.ch}   \quad  \texttt{\small \{kan8,hpirsiav@umbc.edu\}}$}
}
\maketitle
%\thispagestyle{empty}

%%%%%%%%% ABSTRACT
\begin{abstract}
%A remarkable effort has been put on introducing variety of pseudo-tasks to learn high quality visual representation.  However, the methods introduced so far are significantly far a way from supervised learning. We introduce a simple but effective framework to uniformly evaluate and boost the performance of a pseudo-task. Given a pre-trained network on an arbitrary architecture, we cluster images on the feature space of the pre-trained network and train another network with a fixed architecture to predict the cluster-id assigned to each image. We show that this simple framework allows us obtain better labels in clustering by training the original pseudo task on a deeper network. We outperform all the state-of-the-art self-supervised methods on all benchmark with an impressive margin. Our method achieves 56.0\% and 72.0\% on VOC2007 detection and classification, which is closing a big gap between current supervised and self-supervised learning. Moreover, we show that our proposed framework is quite useful when the original pseudo task is trained under different settings, like batch normalization, that prevent from smooth transfer learning.  
In self-supervised learning, one trains a model to solve a so-called pretext task on a dataset without the need for human annotation. The main objective, however, is to transfer this model to a target domain and task. Currently, the most effective transfer strategy is fine-tuning, which restricts one to use the same model or parts thereof for both pretext and target tasks.
In this paper, we present a novel framework for self-supervised learning that overcomes limitations in designing and comparing different tasks, models, and data domains. In particular, our framework decouples the structure of the self-supervised model from the final task-specific fine-tuned model. This allows us to: 1) quantitatively assess previously incompatible models including handcrafted features; 2) show that deeper neural network models can learn better representations from the same pretext task; 3) transfer knowledge learned with a deep model to a shallower one and thus boost its learning. 
We use this framework to design a novel self-supervised task, which achieves state-of-the-art performance on the common benchmarks in PASCAL VOC 2007, ILSVRC12 and Places by a significant margin. Our learned features shrink the mAP gap between models trained via self-supervised learning and supervised learning from $5.9\%$ to $2.6\%$ in object detection on PASCAL VOC 2007. 
\end{abstract}
%%%%%%%%% BODY TEXT
\section{Introduction}

%Novelty:
%Using clustering to unify unsupervised algorithms to do better comparison
%Clustering helps us get a better fine-tuned alexnet model using a deeper self-supervised model
%Occlusion and number of clusters improves puzzle
%Best number out there
%Clustering sometime boost the performance 
%Hierarchical clustering [?]
%We might show that clustering has less dataset bias compared to doing self-supervised learning and fine-tuning on two different datasets. [??]
%
%
%
%
%We have 4 stages:
%Unsupervised learning
%Clustering
%Cluster classification
%Finetuning
%
%For (1), we will use others? features
%For (4), we will use PASCAL (classification, detection, and segmentation)

%pseudo-labels capture the metric induced by a task
%
%VGG can learn more than AN means that the pseudo-labels are different and give better performance in TL

%%%%%%%%%%%%%%%%
%-we are interested in obtaining the best feature possible with self-supervised learning
Self-supervised learning (SSL) has gained considerable popularity since it has been introduced in computer vision \cite{context,colorful,counting,shuffle}. Much of the popularity stems from the fact that SSL methods learn features without using manual annotation by introducing a so-called \emph{pretext task}. %Because no human-annotated labels are used to learn features, SSL is also referred to as  unsupervised learning. However, SSL is not like classic unsupervised learning (such as clustering). In fact, as we explain later, SSL methods may use tools from supervised learning, and learn a mapping from data to artificially created pseudo-labels.
%-every SSL task induces a different metric on the data
Feature representations learned through SSL in computer vision are often transferred to a \emph{target data domain} and a \emph{target task}, such as object classification, detection and semantic segmentation in PASCAL VOC. %\cite{splitBrain}. 
These learned features implicitly define a metric on the data, \ie, which data samples are similar and which ones are dissimilar. Thus, the main objective of a pretext task is to learn a metric that makes images of the same object category similar and images of different categories dissimilar.
%-how do we improve the metric?
A natural question is then: How do we design such a task? 
%-current methods do ...
Some SSL approaches define pretext tasks through explicit desirable invariances of the metric \cite{wangVideo,wangTransitive,DosovitskiyExemplar,counting} or such that they implicitly require a good object representation \cite{context,noroozi2016,colorful}.

\begin{figure}[t]
\begin{center}
\includegraphics[width=1\linewidth,trim=0 1cm 0 0cm,clip]{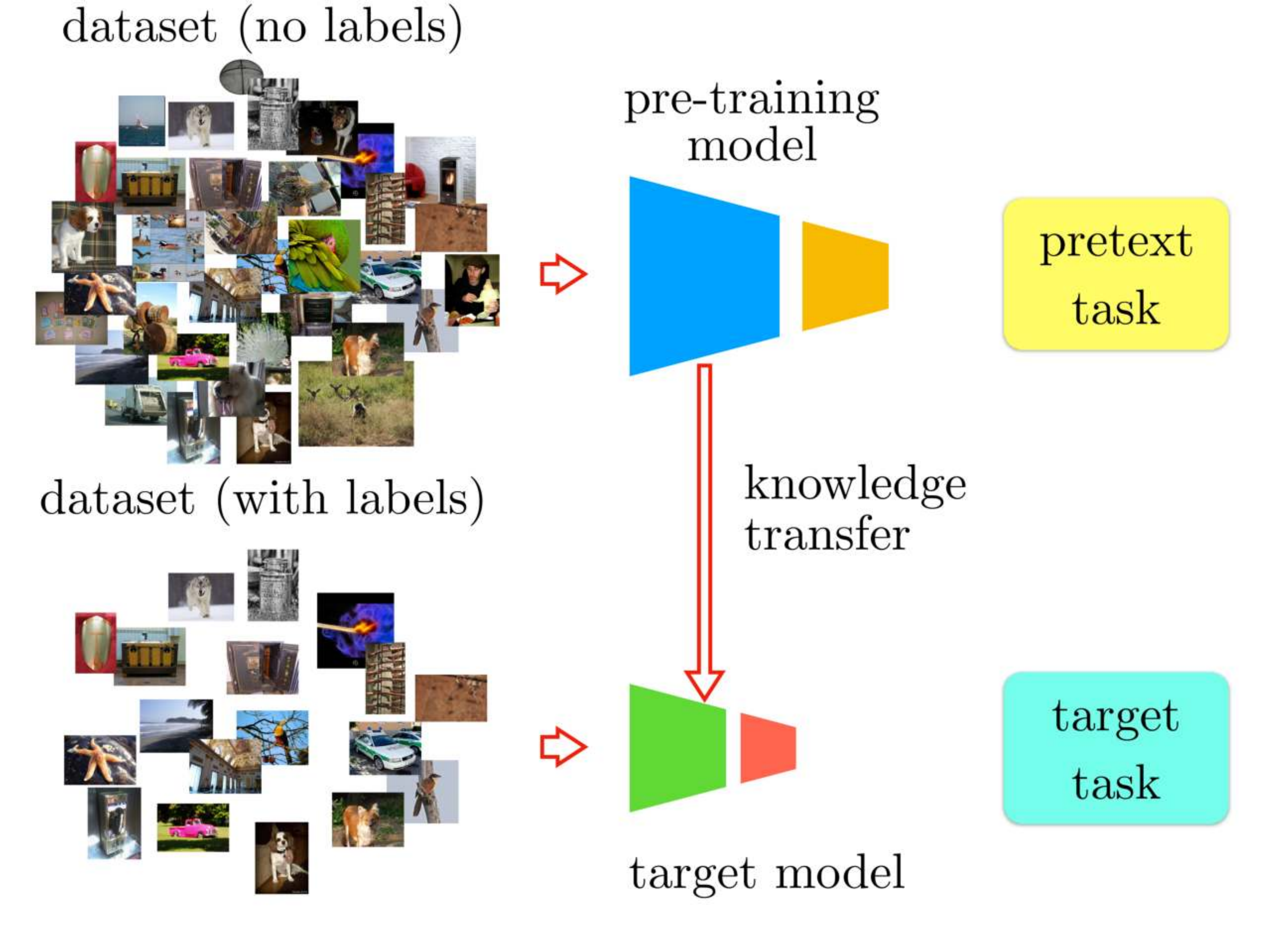}
\end{center}
   \caption{Most current self-supervised learning approaches use the same architecture both in pre-training and fine-tuning. We develop a knowledge transfer method to decouple these two architectures. This allows us to use a deeper model in pre-training.}
\label{fig:challenge}
\end{figure}
%-how do we choose the best?
Even if we had a clear strategy to relate pretext tasks to a target task, comparing and understanding which one is better presents challenges. 
Most of the recent approaches transfer their learned features to a common supervised target task. This step, however, is complicated by the need to use the same model (\eg, AlexNet \cite{AlexNet12}) to solve both tasks. This clearly poses a major limitation on the design choices. For example, some pretext tasks may exploit several data domains (\eg, sound, text, videos), or may exploit different datasets sizes and formats, or might require very deep neural networks to be solved.

%-we observe two main directions:
%-1) we know that difficult but learnable tasks build better representations than simple ones most methods focus on designing new tasks
%-2) we know that deep architectures build better representations than shallow ones some methods use different architectures, but then do not have a simple way to compare them
There is thus the need to build better representations by exploring and comparing difficult, but learnable \cite{SchmidhuberCuriousExplorer}, pretext tasks and arbitrarily deep architectures.
%-to better understand what one can gain through the task or the network design, we propose a transfer learning method so that we can compare (we need to be able to transfer knowledge to the same architecture)
%-transfer knowledge from one Framework to another can be done through distillation: regress features
Towards this goal, one could use methods such as \emph{distillation} \cite{HintonDistillation, caruana} to transfer the representation in a strong model (the one trained with the pretext task) to a smaller one (the one employed on the target task). \hamedrem{These methods, however, require that both models use softmax as the output and with the same dimensionality.}
%-this does not work well (show experiment) and so we discard it
%Another strategy to transfer knowledge from one model to another is to train the target model to predict features computed with the first one. As we show in our experiments, this type of transfer does not work well. 

%-we propose instead to transfer knowledge by reducing the learned representation to pseudo-labels
%how do we get the the pseudo-labels?
%-to obtain the pseudo-labels to propose to cluster the features with k-means

In this paper, we propose to transfer knowledge by reducing a learned representation to \emph{pseudo-labels} on an unlabeled dataset. First, we compute the features learned through a pretext task which might use a complex model on a dataset of unlabeled images. Second, we cluster the features (\eg, using k-means) and use the cluster ID as pseudo-labels for unlabeled images. Third, we learn our final representation by training a smaller deep network (\eg, AlexNet) to classify the images based on the pseudo-labels.
\hamedrem{First, we cluster (via k-means) the features computed by the model trained on the pretext task. Second, we assign the feature corresponding to each data sample to the closest cluster feature. Finally, pseudo-labels of each data sample are the corresponding cluster assignment, and the number of clusters thus defines the number of pseudo-labels.
%We obtain pseudo-labels by clustering (via k-means) the features computed by the model trained on the pretext task. Clustering captures the metric defined by the learned features by assigning a pseudo-label (the corresponding cluster) to each data sample. 
This operation gives a lot of flexibility: The datasets used for clustering, pseudo-label assignment and training on the pretext task can all be different.
}
By reducing the feature representation of a model to data/pseudo-label pairs, it seems that we are discarding a lot of information. 
%Clustering captures the metric defined by the learned features. 
%-this is the same task format that we use with supervised learning, and therefore it potentially  can capture the same representation
\hamedrem{However, notice that this is the same representation used in supervised learning. Therefore, we argue that our knowledge representation can capture the same knowledge we can express with human-annotated labels. Ideally, if we trained a model with supervised learning, the pseudo-labels obtained through feature clustering should match those used during training up to an index permutation. Moreover, we would like a good representation to give good results in a nearest neighbor search, \ie, we would like semantically similar data points to be close to each other. Hence, pseudo-labels obtained through a simple clustering algorithm should be a robust estimator of the learned representation.}
%-moreover: A good representation should have good results in nearest neighbor search i.e., semantically similar points should be close to each other. Hence, a simple clustering algorithm should group semantically similar data-points.
%\hamed{This is true, but we believe reducing the problem to standard supervised classification setting, lets us use 
However, we know that given good labels that group semantically similar images, standard supervised classification methods work well. Also, we believe that in a good representation space, semantically related images should be close to each other. Hence, pseudo-labels obtained through a simple clustering algorithm should be a robust estimate of the learned representation.

%-then the transfer of knowledge between networks becomes a classification task on the pseudo-labels
Once we have obtained pseudo-labels on some dataset, we can \emph{transfer knowledge by simply training a model to predict those pseudo-labels}.
% -with our transfer technique we now can: 
%1) regardless of the original arch, its settings and training we can transfer knowledge to a fixed arch (this allows to capture representation qualities of deeper architectures, as well as more of the complexity of the source tasks);
%2) we can compare different models (network architectures or handcrafted features) built on different data domains and with different training tasks, by using a common reference model, data and task.
The simplicity and flexibility of our technique allows us to: 
\begin{enumerate}
\item Transfer knowledge from any model (different network architecture and training settings) to any other final task model; we can thus capture representations of complex architectures as well as complicated pretext tasks (see Figure~\ref{fig:challenge}).

\item Compare different models (\eg, learned features via neural network architectures versus handcrafted features), built on different data domains with different pretext tasks using a common reference model, data, and task (e.g., AlexNet, PASCAL VOC, and object detection)
\end{enumerate}
%-Based on this analysis we design a better self-supervised task.
%-We build on the Puzzle method, add occlusions and more permutations
%-We boost its performance just by training it on VGG and then transferring it to AlexNet (pseudo-labels)
Based on this analysis, to show the effectiveness of our algorithm, we design a novel self-supervised task that is more complicated and uses a deeper model. We start from an existing pretext task, the jigsaw problem \cite{noroozi2016}, and make it more challenging by adding occlusions. We refer to this task as the \emph{Jigsaw++} problem. Furthermore, we boost its performance by training it on VGG16 \cite{VGG} and then transferring to AlexNet \cite{AlexNet12} via our proposed transfer method. The resulting model achieves state-of-the-art-performance on several benchmarks shrinking the gap with supervised learning significantly. Particularly, on object detection with Fast R-CNN on PASCAL VOC 2007, Jigsaw++ achieves $56.5\%$ mAP, while supervised pre-training on ImageNet achieves $59.1\%$ mAP. Note that the final model in both cases uses the same AlexNet architecture. 
We believe our knowledge transfer method can potentially boost the performance of shallow representations with the help of more complicated pretext tasks and architectures.

\hamedrem{We expect that by training Jigsaw++ on deeper networks such as ResNet \cite{ResNet} and with a larger dataset, the performance of AlexNet can be further boosted and perhaps surpass that of supervised learning pre-training.}

\section{Prior Work}
\noindent\textbf{Self-supervised learning.} 
As mentioned in the introduction, SSL consists of learning features using a pretext task. %This task provides implicitly or explicitly a \emph{supervision signal}. We provide an overview of SSL methods based on the  can categorize a self-supervised method based on the source of the supervision signal that it uses to design the pseudo-task. 
Some methods define as task the reconstruction of data at the pixel level from partial observations and are thus related to denoising autoencoders \cite{DAE06}. Notable examples are the \emph{colorization problem} \cite{colorful,larsson2016learning}, where the task is to reconstruct a color image given its gray scale version. Another example is image inpainting \cite{ContextEncoder}, where the task is to predict a region of the image given the surrounding. Another category of methods uses temporal information in videos. Wang and Gupta \cite{wangVideo} learn a similarity metric using the fact that tracked patches in a video should be semantically related.
Another type of pretext task, is the reconstruction of more compact signals extracted from the data. For example, Misra \etal and Brattoli \etal~\cite{shuffle, buechlerCVPR17} train a model to discover the correct order of video frames. 
Doersch \etal~\cite{context} train a model that predicts the spatial relation between image patches of the image. Noroozi and Favaro \cite{noroozi2016} propose to solve jigsaw puzzles as a pretext task. 
Pathak \etal~\cite{WatchingObjects} obtain a supervisory signal by segmenting an image into foreground and background through the optical flow between neighboring frames of a video. Then, they train a model to predict this segmentation from a single image.
Other methods also use external signals that may come freely with visual data. The key idea is to relate images to information in other data domains like ego-motion \cite{tiedEgomotion, agrawal15} or sound \cite{ambientSound}. Finally, recent work \cite{counting} has also exploited the link between relative transformations of images to define relative transformations of features.

%Despite the large effort on designing self-supervised learning tasks, they are still far from supervised learning in transfer learning benchmarks. Recently, mulit-task leanring a
Currently, the above pretext tasks have been assessed through transfer learning and benchmarked on common neural network architectures (\eg, AlexNet) and datasets (\eg, PASCAL). However, so far it is unclear how to design or how to further improve the performance of SSL methods. One natural direction is the combination of multiple tasks \cite{doersch2017multi,wangTransitive}. However, this strategy does not seem to scale well as it becomes quickly demanding in terms of computational resources. Moreover, a possible impediment to progress is the requirement of using the same model both for training on the pretext task and to transfer to another task/domain. In fact, as we show in our experiments, one can learn better representations from challenging tasks by using deep models, than by using shallower ones. Through our knowledge transfer method we map the representation in the deep model to a reference model (\eg, AlexNet) and show that it is better than the representation learned directly with the reference model. This allows to improve SSL methods by exploring: 1) designs of architectures that may be more suitable for learning a specific pretext task; 2) data formats and types different from the target domain; 3) more challenging pretext tasks.
To illustrate these advantages, we make the method of Noroozi and Favaro \cite{noroozi2016} more challenging by incorporating occlusions in the tiles.
%All of these pretext tasks have different levels of complexity and capture different representations. A more complex but learnable task will potentially lead to the better representations. However, it needs more complex model to be solved. This makes the transfer learning more challenging. We introduce the jigsaw++  which is a complex but not ambiguous task and also introduce a framework that allows us to solve the pseudo-task in a complex model and transfer the knowledge results from solving the task to a shallow model.

\noindent\textbf{Knowledge distillation.} Since we transfer knowledge from a model trained on a pretext task to a target model, our work is related to \emph{model distillation}. \cite{HintonDistillation,DistillationCaruana} perform knowledge distillation by training a target model that mimics the output probability distribution of the source model. \cite{Wang-2016-4848,attensionTransfer} extend that method to regressing neuron activations, an approach that is more suitable to our case. Our approach is fundamentally different. We are only interested in preserving the essential metric of the learned representation (the cluster associations), rather than regressing the exact activations. %Hence, we transfer the knowledge using a clustering algorithm. 
\hamedrem{There are two main approaches to distill a model. The first one is \emph{knowledge distillation} \cite{HintonDistillation, DistillationCaruana}. The idea is to approximate the distribution of the representation of a \emph{teacher model} with a \emph{student model} by minimizing a softmax-based loss. The second approach is \emph{attention transfer} \cite{attensionTransfer}, where the student model mimics the teacher model layer-wise through a fixed transformation. Both approaches are challenging to train in a general setting because approximating a high-dimensional probability distribution is challenging and requires engineering of both teacher and student models. Our proposed knowledge transfer %model distillation 
approach is simple, efficient, and robust. By using clustering we capture the structure of the feature space in a quite concise representation (pseudo-labels), which would reveal if semantically similar data points have features close to each other.}
A clustering technique used in Dosovitskiy \etal~\cite{DosovitskiyExemplar} is very related to our method. They also use clustering to reduce the classification ambiguity in their task, when too many surrogate classes are used. However, they only train and re-train the same network and do not exploit it for knowledge transfer. Other related work uses clustering in the feature space of a supervised task to extract labels from unlabeled data for novel categories~\cite{R2A1_NIPS2016_6408} or building hash functions~\cite{R2A2_hashlearning}. Neither of these works uses clustering to transfer knowledge from a deep network to a shallow one. Our HOG experiment is related to \cite{cimpoi2016deep} which shows the initial layers of VGG perform similarly to hand-crafted SIFT features.
%consistent with the nature of self-supervised representation learning, where we have already a representation space that embeds semantic. Hence, it builds a good similarity metric to group data-points.

%\section{Self-Supervised Learning}

%-we are interested in obtaining the best feature possible with self-supervised learning
%
%-every SSL task induces a different metric on the data
%
%-how do we choose the best/how do we improve the metric?
%
%-current methods do ... (see Hamed's)
%
%-we observe two main directions:
%
%-1) we know that difficult but learnable tasks build better representations than simple ones
%most methods focus on designing new tasks
%
%-2) we know that deep architectures build better representations than shallow ones
%some methods use different architectures, but then do not have a simple way to compare them

\section{Transferring Knowledge}

The common practice in SSL is to learn a rich representation by training a model on a SSL pretext task with a large scale unlabeled dataset and then fine-tune it for a final supervised task (\eg, PASCAL object detection) by using a limited amount of labeled training data. This framework has an inherent limitation: The final task model and the SSL model must use the same architecture. This limitation becomes more important as the community moves on to more sophisticated SSL pretext tasks with larger scale datasets that need more complicated deeper model architectures. Since we do not want to change the architecture of the final supervised task, we need to develop a novel way of transferring the learned knowledge from the SSL task to the final supervised model. Moreover, when trained on the pretext task, the model learns some extra knowledge that should not be transferred to the final task. For instance, in standard fine-tuning, we usually copy the weights only up to some intermediate convolutional layers and ignore the final layers since they are very specific to the pretext task and are not useful for general visual recognition. 

\begin{figure}[t]
\begin{center}
\renewcommand{\arraystretch}{1.5}% Spread rows out...
\begin{tabular}{@{}>{\centering}m{0in} >{\centering\arraybackslash}m{1\linewidth}}
(a) &
\includegraphics[width=.9\linewidth,trim={0 4cm 0cm 0cm},clip]{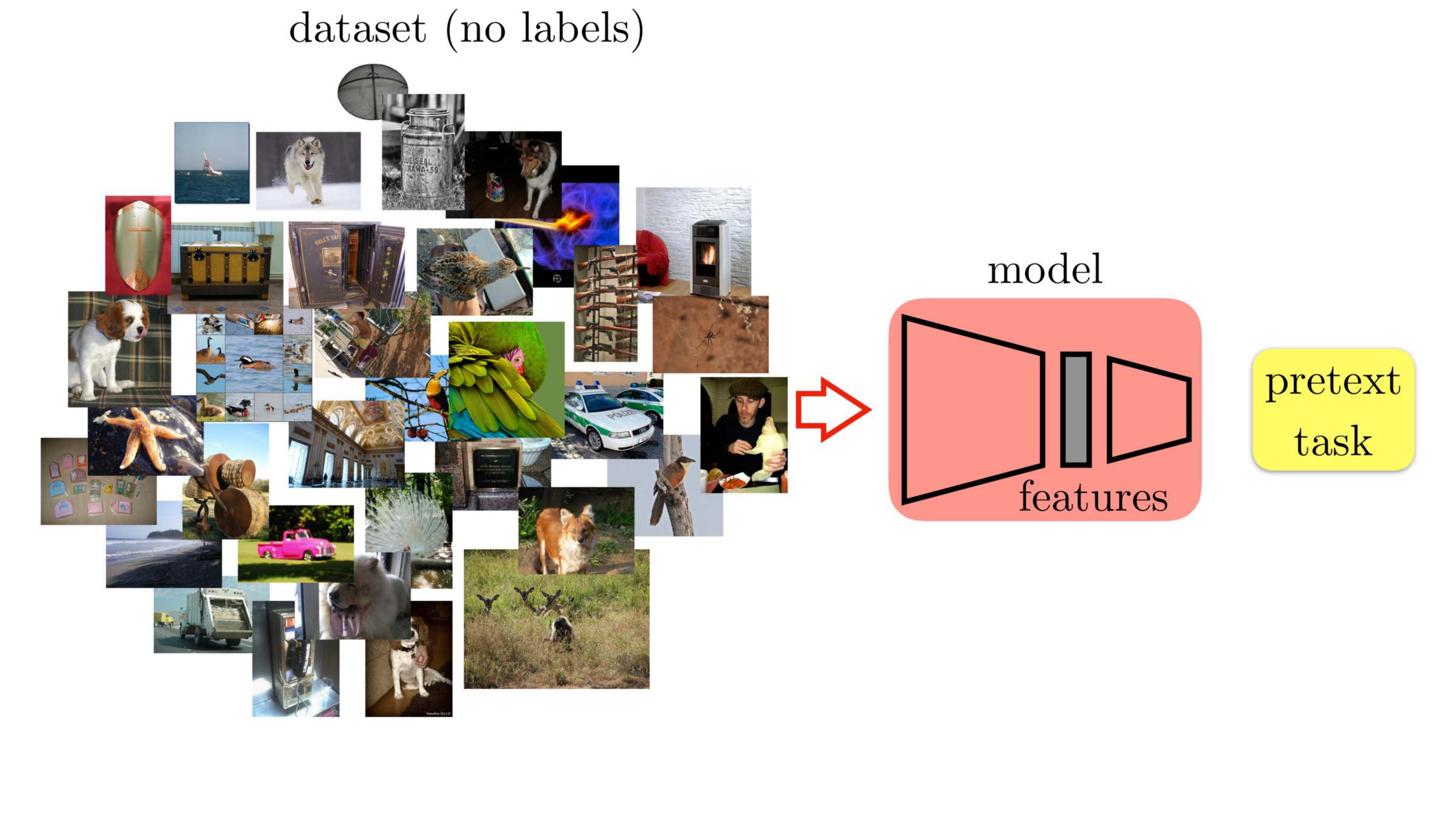}\\
(b) &
\includegraphics[width=.9\linewidth,trim={0 9cm 0cm 0cm},clip]{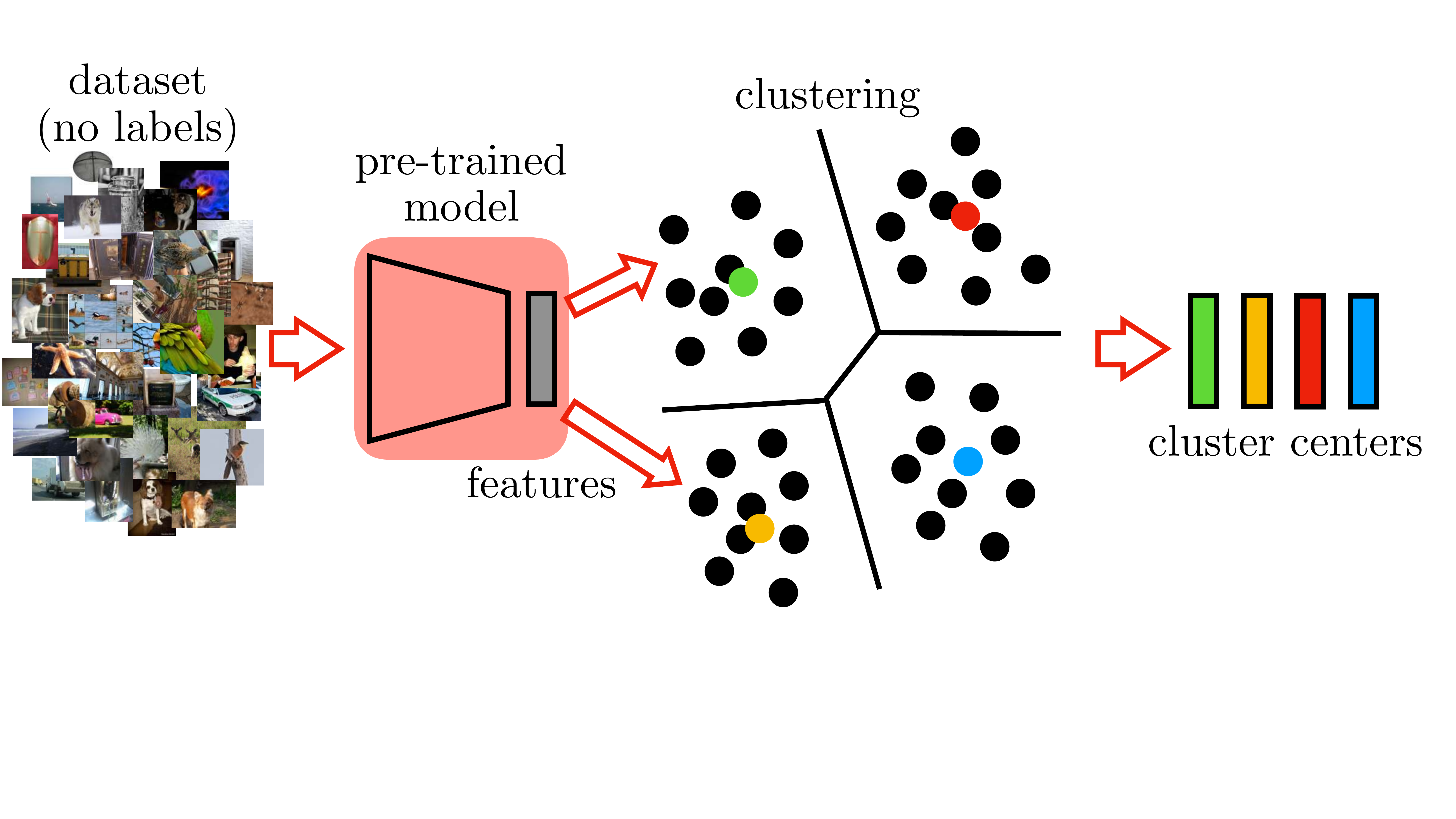}\\
(c) &
\includegraphics[width=.9\linewidth,trim={0 9cm 0cm 0cm},clip]{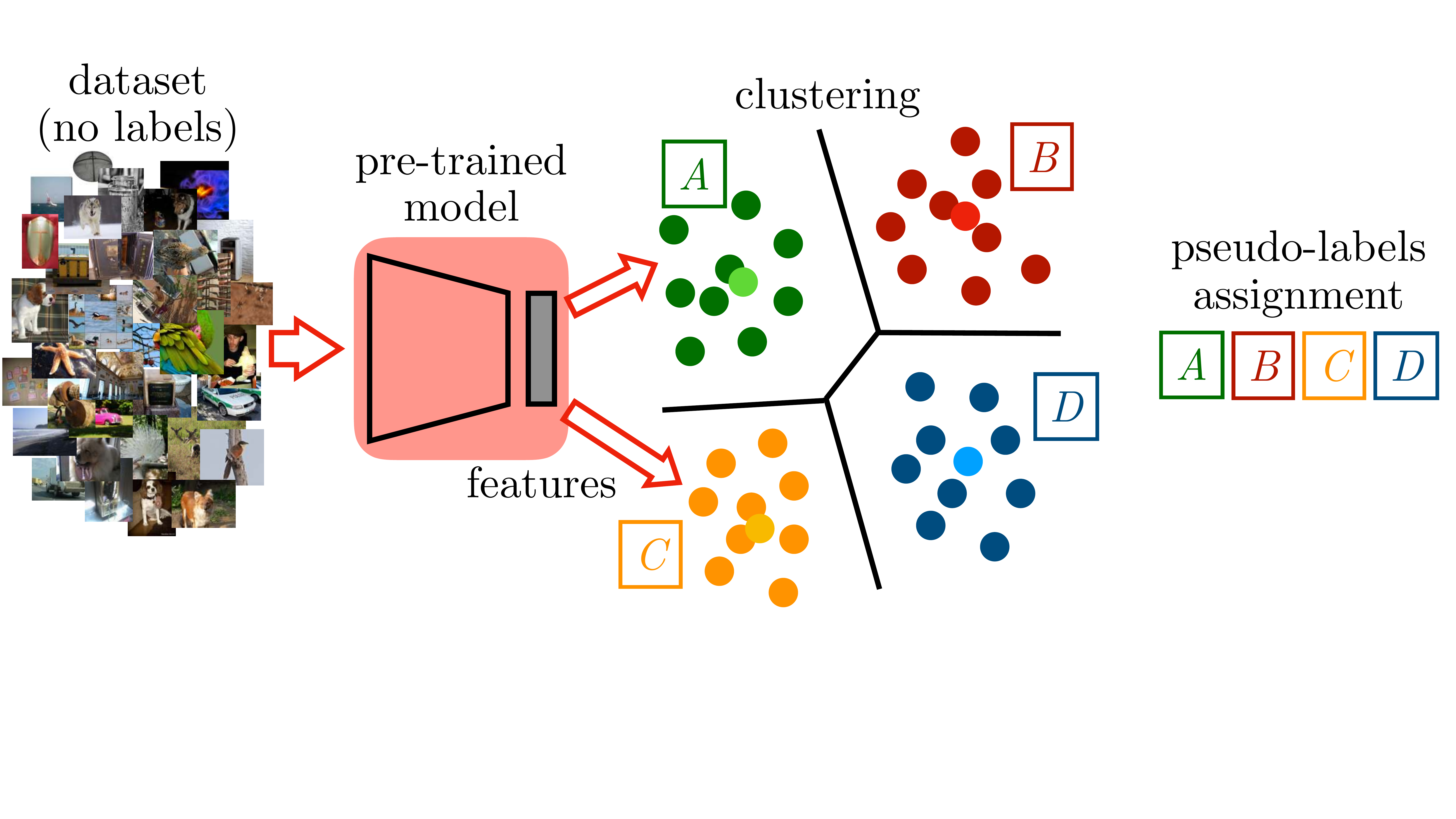}\\
(d) &
\includegraphics[width=.9\linewidth,trim={0 4cm 0cm 0cm},clip]{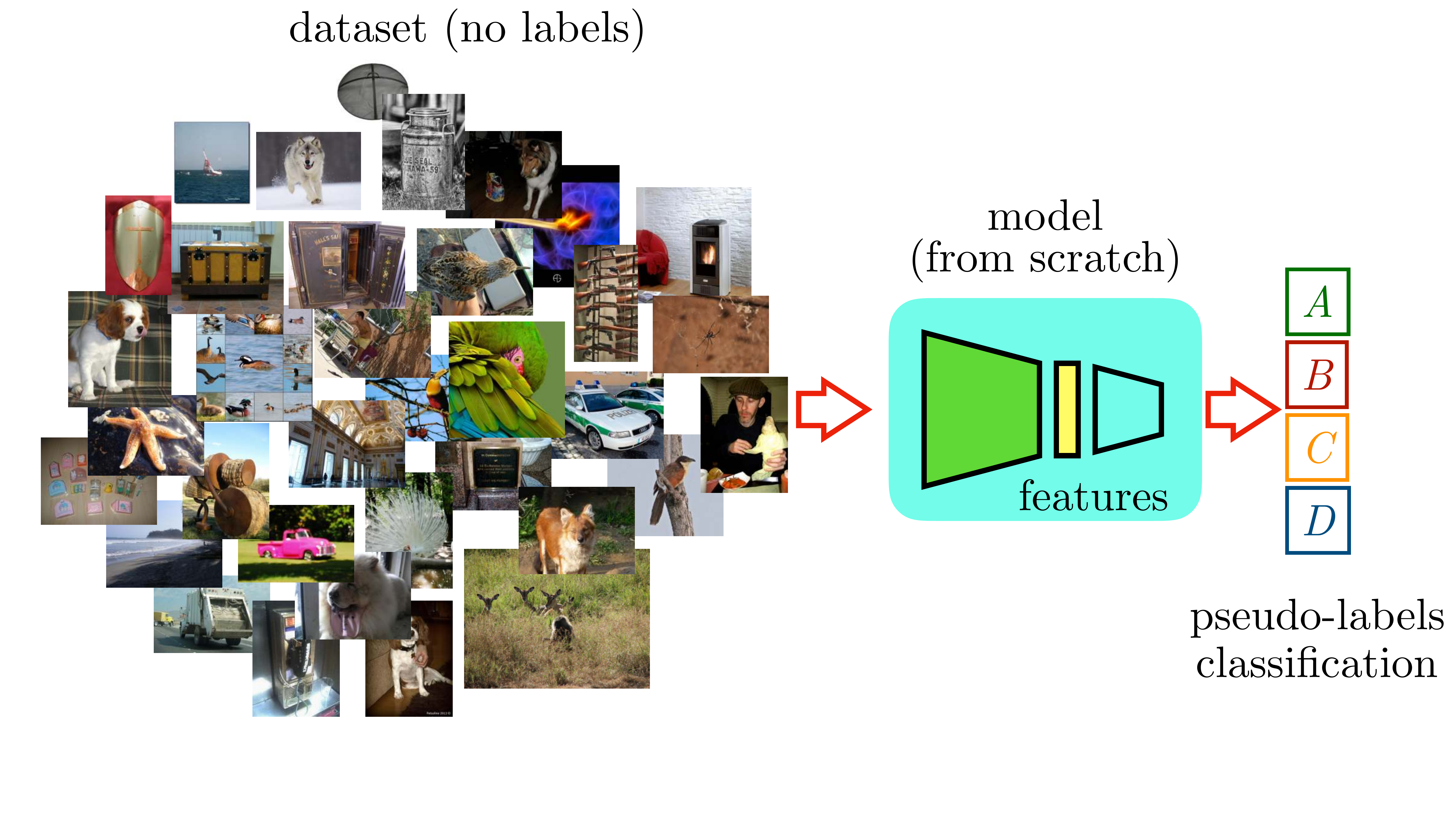}
\end{tabular}
\end{center}
   \caption{\textbf{Knowledge transfer pipeline.} We break down the four steps of our proposed method for knowledge transfer: (a) an arbitrary model is pre-trained on an SSL pretext task; (b) the features extracted from this model are clustered and cluster centers are extracted; (c) pseudo-labels are defined for each image in the dataset by finding the closest cluster center; (d) training of the target model on the classification of the pseudo-labels.}
\label{fig:pipeline}
\end{figure}

In this section, we propose an algorithm to transfer the part of knowledge learned in SSL that is useful for visual recognition to the target task. Our idea is based on the intuition that in the space of a good visual representation, semantically similar data points should be close to each other. The common practice to evaluate this is to search for nearest neighbors and make sure that all retrieved results are semantically related to the query image. This means a simple clustering algorithm based on the Euclidean distance should group semantically similar images in the same cluster. Our idea is to perform this clustering in the feature space and to obtain the cluster assignments of each image in the dataset as pseudo-labels. We then train a classifier network with the target task architecture on the pseudo-labels to learn a novel representation. We illustrate our pipeline in Figure~\ref{fig:pipeline} and describe it here below.\\
\noindent\textbf{(a) Self-Supervised Learning Pre-Training.}
Suppose that we are given a pretext task, a model and a dataset. Our first step in SSL is to train our model on the pretext task with the given dataset (see Figure~\ref{fig:pipeline}~(a)).
Typically, the models of choice are convolutional neural networks, and one considers as feature the output of some intermediate layer (shown as a grey rectangle in Figure~\ref{fig:pipeline}~(a)).
%After training the network weights are frozen.\\

\noindent\textbf{(b) Clustering.}
Our next step is to compute feature vectors for all the unlabeled images in our dataset. Then, we use the k-means algorithm with the Euclidean distance to cluster the features (see Figure~\ref{fig:pipeline} (b)). Ideally, when performing this clustering on ImageNet images, we want the cluster centers to be aligned with object categories. In the experiments, we typically use 2{,}000 clusters.

\noindent\textbf{(c) Extracting Pseudo-Labels.}
The cluster centers computed in the previous section can be considered as \emph{virtual categories}. Indeed, we can assign feature vectors to the closest cluster center to determine a \emph{pseudo-label} associated to the chosen cluster. This operation is illustrated in Figure~\ref{fig:pipeline} (c). Notice that the dataset used in this operation might be different from that used in the clustering step or in the SSL pre-training.

\noindent\textbf{(d) Cluster Classification.}
Finally, we train a simple classifier using the architecture of the target task so that, given an input image (from the dataset used to extract the pseudo-labels), predicts the corresponding pseudo-label (see Figure~\ref{fig:pipeline} (d)). This classifier learns a new representation in the target architecture that maps images that were originally close to each other in the pre-trained feature space to close points.

\begin{figure}[t]
\begin{center}
\begin{subfigure}[]{.22\textwidth}
\includegraphics[width=1\linewidth,clip]{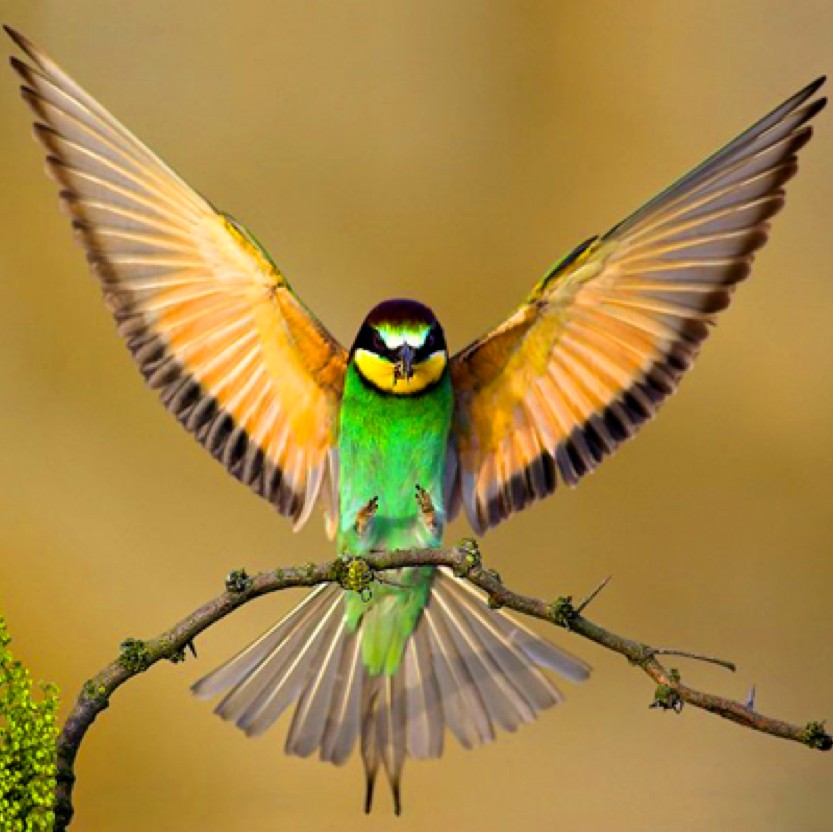}
\vspace{-.6cm}
\caption{}
\end{subfigure}\hspace{.4cm}
\begin{subfigure}[]{.22\textwidth}
\includegraphics[width=1\linewidth,clip]{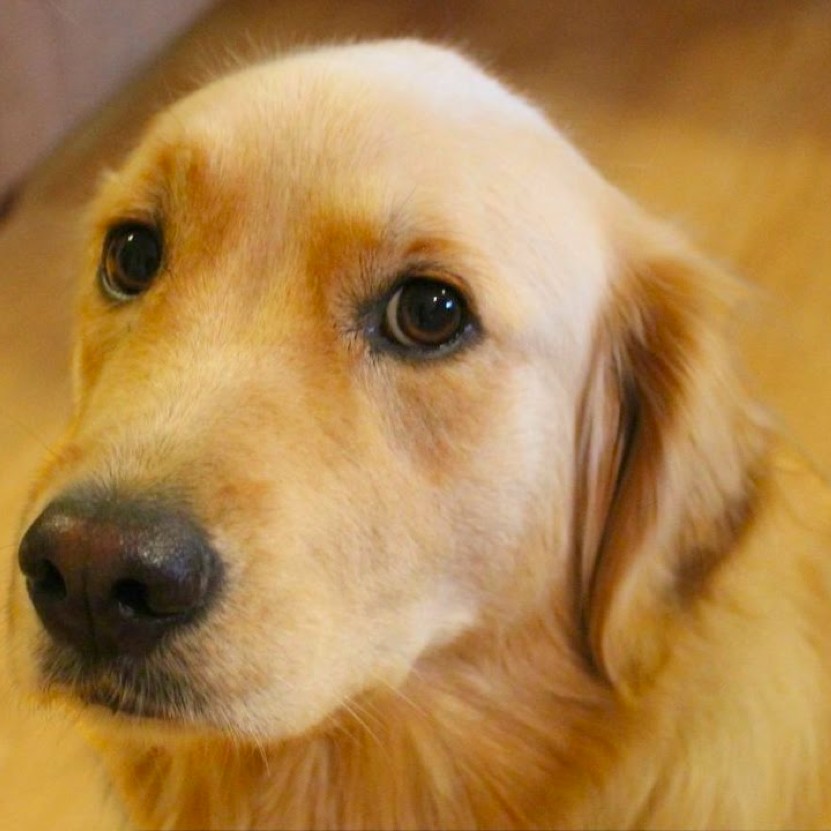}
\vspace{-.6cm}
\caption{}
\end{subfigure}
\begin{subfigure}[]{.22\textwidth}
\includegraphics[width=1\linewidth,trim={5cm 0cm 5cm 0cm},clip]{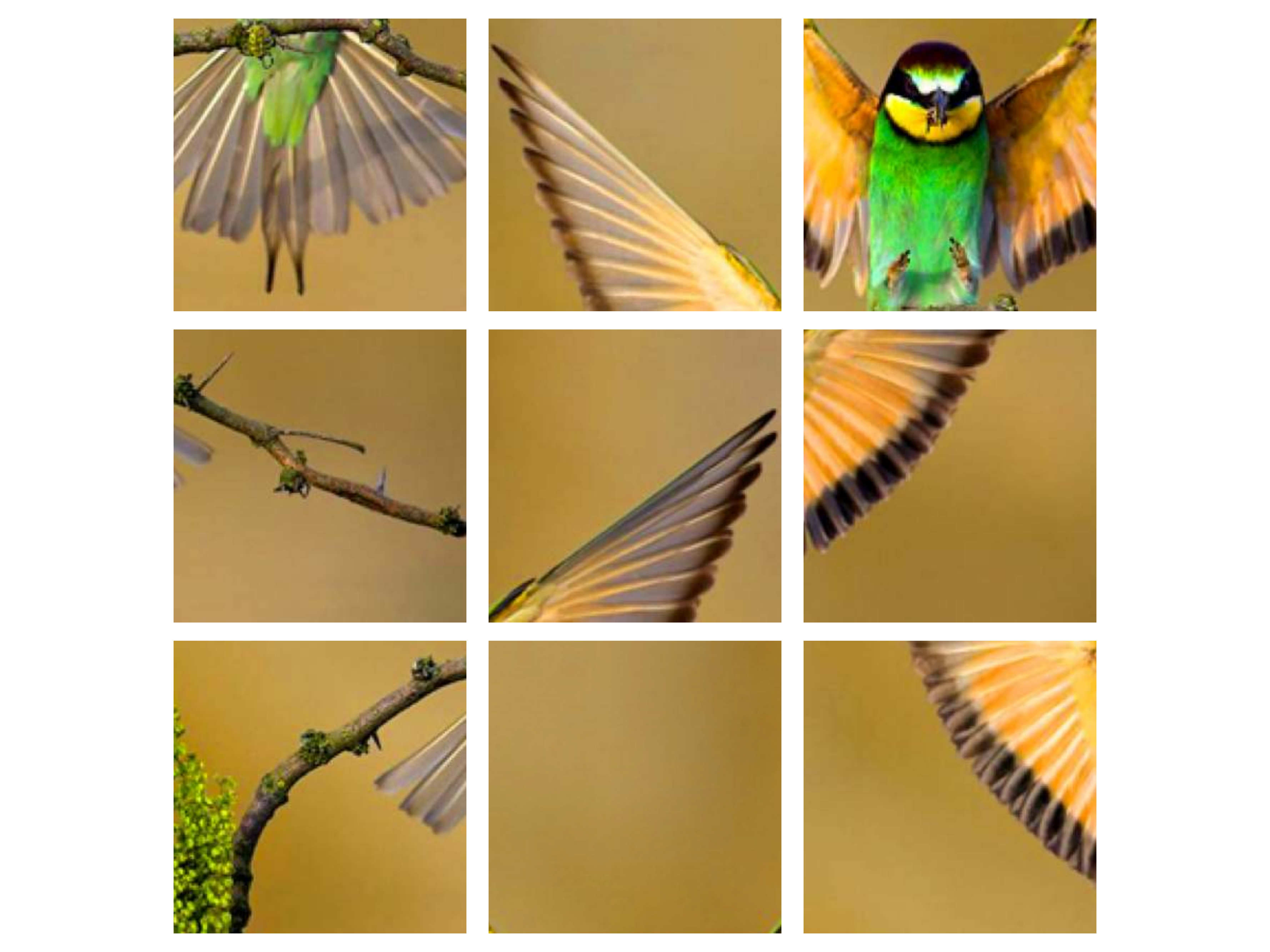}
\vspace{-.7cm}
\caption{}
\end{subfigure}\hspace{.4cm}
\begin{subfigure}[]{.22\textwidth}
\includegraphics[width=1\linewidth,trim={5cm 0cm 5cm 0cm},clip]{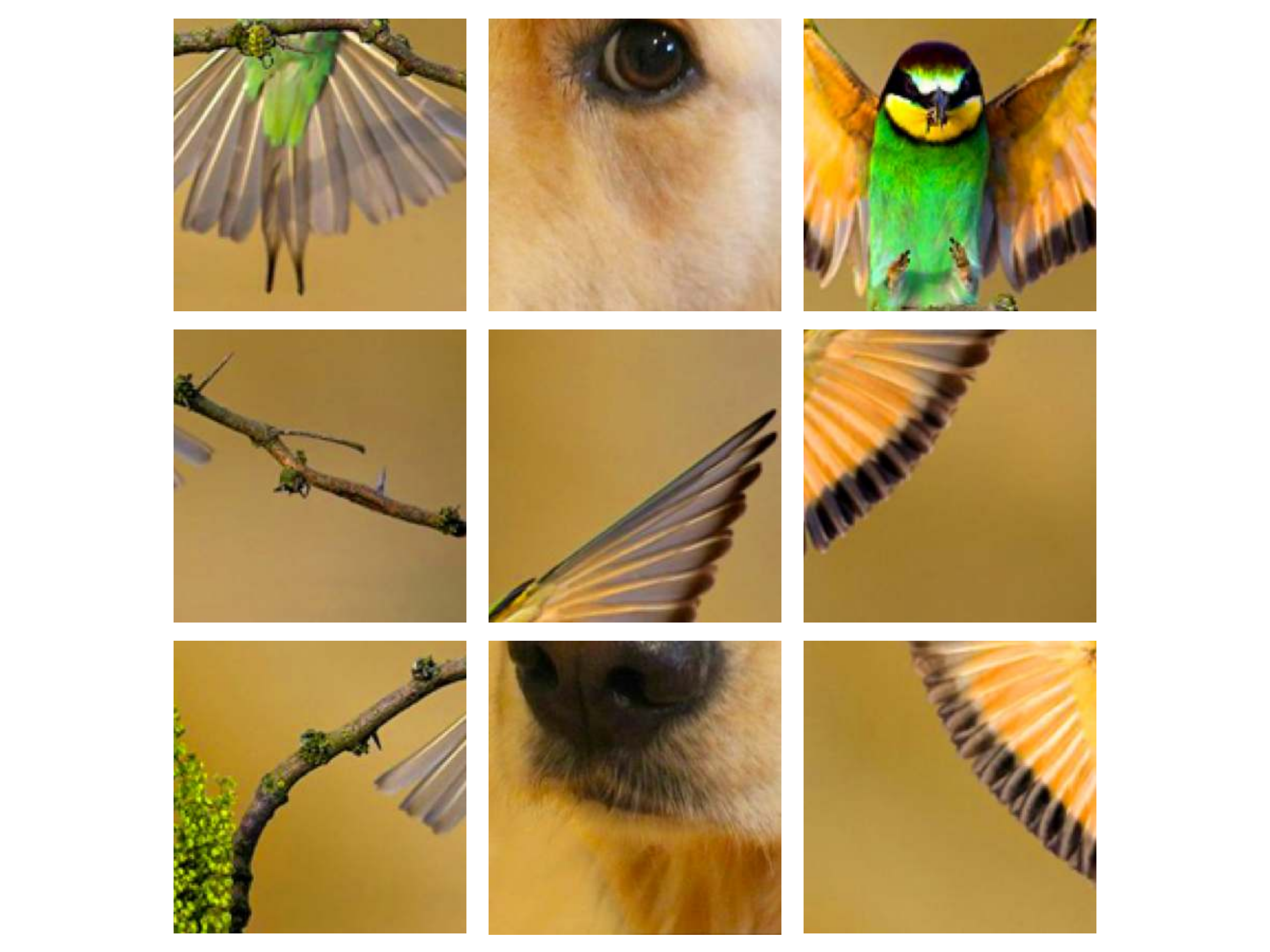}
\vspace{-.7cm}
\caption{}
\end{subfigure}
\vspace{-.5cm}
\end{center}
   \caption{\textbf{The Jigsaw++ task.} (a) the main image. (b) a random image. (c) a puzzle from the original formulation of \cite{noroozi2016}, where all tiles come from the same image. (d) a puzzle in the Jigsaw++ task, where at most $2$ tiles can come from a random image.}
\label{fig:jigsaw}
\end{figure}

\section{The Jigsaw++ Pretext Task}
\label{sec:jigsaw}

Recent work \cite{doersch2017multi,wangTransitive} has shown that deeper architectures can help in SSL with PASCAL recognition tasks (\eg, ResNet). However, those methods use the same deep architecture for both SSL and fine-tuning. Hence, they are not comparable with previous methods that use a simpler AlexNet architecture in fine-tuning. We are interested in knowing how far one can improve the SSL pre-training of AlexNet for PASCAL tasks. Since in our framework the SSL task is not restricted to use the same architecture as in the final supervised task, we can increase the difficulty of the SSL task along with the capacity of the architecture and still use AlexNet at the fine-tuning stage.

Towards this goal, we build on the method of Okanohara \etal~\cite{Okanohara07} to learn representations in the text domain. They replace a word at random in a sentence and train a model to distinguish the original sentence from the corrupt one. We combine this idea with the jigsaw \cite{noroozi2016} task by replacing tiles in the image puzzle with a random tile from other images. %This approach would not achieve a high quality representation space as the visual data is not segmented as text data. We propose to employ this idea accompanied by the jigsaw puzzle task.
We call this the \emph{Jigsaw++ task}.
The original pretext task \cite{noroozi2016} is to find a reordering of tiles from a $3\times 3$ grid of a square region cropped from an image.  In Jigsaw++, we replace a random number of tiles in the grid (up to $2$) with (occluding) tiles from another random image (see Figure~\ref{fig:jigsaw}). The number of occluding tiles (0, 1 or 2 in our experiments) as well as their location are randomly selected. The occluding tiles make the task remarkably more complex. First, the model needs to detect the occluding tiles and second, it needs to solve the jigsaw problem by using only the remaining patches. To make sure we are not adding ambiguities to the task, we remove similar permutations so that the minimum Hamming distance between any two permutations is at least $3$. In this way, there is a unique solution to the jigsaw task for any number of occlusions in our training setting. Our final training permutation set includes $701$ permutations, in which the average and minimum Hamming distance is $.86$ and $3$ respectively. In addition to applying the mean and std normalization independently at each image tile, we train the network $70\%$ of the time on gray scale images. In this way, we prevent the network from using low level statistics to detect occlusions and solve the jigsaw task. 
\begin{table}[t]
\caption{Impact of the number of cluster centers}
\label{tab:clusters}
\footnotesize
\centering
%\begin{adjustbox}{width=.48\textwidth}
\begin{tabular}{@{}l@{\hspace{2.2em}}  c@{\hspace{1.5em}}  c@{\hspace{1.5em}}   c@{\hspace{1.5em}} c@{\hspace{1.5em}} c@{}}
\toprule%\hline
\textbf{\#clusters}   & \textbf{500} & \textbf{1000} & \textbf{2000} & \textbf{5000} & \textbf{10000} \\
\textbf{mAP on voc-classification} & 69.1 & 69.5 & 69.9 & 69.9 & 70.0 \\
\bottomrule%\hline
\end{tabular}
\end{table}
We train the Jigsaw++ task on both VGG16 and AlexNet architectures. By having a larger capacity with VGG16, the network is better equipped to handle the increased complexity of the Jigsaw++ task and is capable of extracting better representations from the data. 
Following our pipeline in Figure~\ref{fig:pipeline}, we train our models with this new SSL task, transfer the knowledge by: 1) clustering the features, 2) assigning pseudo-labels, and 3) training AlexNet to classify the pseudo-labels. We execute the whole pipeline by training VGG16 and AlexNet on the Jigsaw++ task. Our experiments show that when we train VGG16 with the Jigsaw++ task, there is a significantly better performance in fine-tuning. This confirms that training on a deeper network leads to a better representation and corresponding pseudo-labels. %Note that in the end we are evaluating the the representations of the cluster classification network which is based on AlexNet architecture in both cases.

%Show 1 picture with example puzzle + occluded tiles

%-------------------------------------------------------------------------
\begin{table}[t]
\caption{Impact of the data domain used in clustering and pseudo-labels: We perform experiments where the training of clustering and extracting pseudo-labels (inference of clustering) are done on different datasets. We see just a little reduction in VOC2007 classification which means the clustering is not relying on the ImageNet bias. }
\label{tab:domain}
\footnotesize
\centering
%\begin{adjustbox}{width=.48\textwidth}
\begin{tabular}{@{} l @{\hspace{2.8em}}  c  c  c }
\toprule%\hline
\textbf{clustering on}   & ImageNet & ImageNet & Places\\
\textbf{pseudo-labels on}   & ImageNet & Places & ImageNet\\
\hline
\textbf{mAP on voc-classification} & 69.9 & 68.4 & 68.3 \\
\bottomrule%\hline
\end{tabular}
\end{table}

\section{Experiments}

We extensively evaluate the knowledge transfer method and the Jigsaw++ task on several transfer learning benchmarks including: fine-tuning on PASCAL VOC, nonlinear classification on ImageNet, and linear classification on Places and ImageNet. We also perform ablation studies to show the effect of the number of clusters and the datasets used for clustering and pseudo-label assignment in the transferred knowledge. 
Our experiments show that pre-training on the Jigsaw++ task yields features that outperform current self-supervised learning methods. 
%As described in \cite{noroozi2016}, the first layer stride of AlexNet is set to $2$ during training and is then set to $4$ as in the standard AlexNet, when transferring to the target task. 
In all the evaluations, the weights of the convolutional layers of the target AlexNet model are copied from the corresponding layers of the AlexNet model trained on the pretext task, and the fully connected layers are randomly initialized.
We evaluate our knowledge transfer method in several cases while the cluster classification network is always AlexNet:\\
\noindent{\bf 1) Jigsaw++ trained with VGG16:} This procedure achieves the best performance on all the benchmarks.\\
\noindent{\bf 2) Jigsaw++ trained with AlexNet:} Our experiments show that the performance does not drop in this case. This implies that a pretext task can be reduced to a classification task, if it has learned a proper similarity metric.\\
\noindent{\bf 3) The method of Doersch \etal ~\cite{context} trained on AlexNet:} Due to the difficulty of their task, they use batch normalization~\cite{batch_norm} during training. This makes the fine-tuning challenging, because of the difference in the settings of the target framework. To address this issue, Kr\"ahenb\"uhl \etal~\cite{krahenbuhl2015data} rescale the weights and thus boost the performance in fine-tuning. Our experiments show that our pipeline is doing significantly better than ~\cite{krahenbuhl2015data} in this case.\\
\noindent{\bf 4) HOG:} We cluster the HOG features to obtain the pseudo-labels. Surprisingly, HOG pseudo-labels yield a high performance in fine-tuning on classification in Pascal VOC.\\
%which is better than any SSL algorithm published before 2018.\\%, compared to the current SSL methods.
%We transfer the HOG features to AlexNet and perform all the  In these experiments, we evaluate the \emph{jigsaw++} task described in Sec.~\ref{sec:jigsaw}. 
%We also perform an extensive evaluation of knowledge transfer of some SSL methods and the jigsaw++ task on the common benchmarks PASCAL VOC 2007/2012, ILSVRC12 and Places \cite{places}.
%These evaluations use several transfer learning protocols: fine-tuning, linear classifiers (on different frozen layers), and nonlinear classifiers (on different frozen layers).
\noindent{\bf 5) Pseudo-labels of a random network:} Zhang \etal~\cite{generalization} showed that AlexNet can be trained to predict random labels. We perform an experiment in the same spirit and obtain random pseudo-labels by clustering the features of a randomly initialized network. To make the cluster classification network converge in this case, we decreased the initial learning rate to $0.001$. By using this settings, the network is able to predict random pseudo-labels on the training set with a high accuracy. However, its prediction accuracy on the validation set is close to random chance as expected. 
%Zhang \etal~\cite{generalization} also carry out a similar experiment and mention that training AlexNet with weight decay fails to converge on random labels. %However, it turned out that it is able to converge on random pseudo-labels that might be of higher quality than random labels, they result from clustering of random edge projection. 
We found that the classifier network trained on random pseudo-labels yields the same performance on transfer learning on PASCAL VOC as random initialization.\\ %The first layer filters of this network is shown in Figure~\ref{fig:filters} which looks better than random filters.
\noindent\textbf{6) Knowledge distillation:} The standard knowledge distillation~\cite{HintonDistillation} is not directly applicable in many SSL tasks
where the loss function does not involve any probability
distribution. Similar to~\cite{DistillationCaruana}, we train a student network
to regress conv4-5 layers of the teacher network (both
AlexNet). The student gets 58.5\% on VOC classification
while the teacher gets 69.8\%. We compare our method to~\cite{attensionTransfer} that minimizes the original loss function as well in distillation. We use the Jigsaw++ network trained on VGG as the teacher network and minimize eq.~(5) of~\cite{attensionTransfer} on AlexNet (the
student), where $\mathcal{L}$ has been replaced by the Jigsaw++ loss. As it is shown in  Table~\ref{tbl:voc}, this method does not boost sensibly the performance on Pascal-VOC dataset.\\
%The student achieves 70.6\% in fine-tuning for VOC classification.
%This shows an improvement of 0.8\% compared
%to training jigsaw++ solely on AlexNet (69.8\%). Our proposed
%method (72.5\%) outperforms both these models.}
%It is challenging to adapt standard knowledge distilation methods for our applicaion. Inspired by \cite{Wang-2016-4848,attensionTransfer}, we train the target network to regress the pre-trained features. Regressing the final layer only results in a very low performance on transfer learning. Hence, we regress all 5 convolutional layers with an $l_2$ loss. This model achieves $58.5\%$ on VOC classification task.\\
\noindent\textbf{Implementation Details.} We extract the features for all the methods from \texttt{conv4} layer and max-pool them to $5\times 5 \times 384$ in the case of AlexNet and $4\times 4 \times 512$ in the case of VGG16. We implement the standard k-means algorithm on a GPU with $k=2K$. It takes around $4$ hours to cluster the $1.3M$ images of ImageNet on a single Titan X. We use the standard AlexNet and ImageNet classification settings to train the pseudo-label classifier network.

\begin{table}[t]
%\begin{center}
 \caption{{\textbf{PASCAL VOC fine-tuning}. Evaluation of SSL methods on classification, detection and semantic segmentation. All rows use the AlexNet architecture at fine-tuning. ``CC+'' stands for ``cluster classification'', our knowledge transfer method. ``vgg-'' means the VGG16 architecture is used before the ``CC+'' step. In ``CC+vgg-Jigsaw++'', we train Jigsaw++ using VGG16, cluster, and train AlexNet to predict the clusters. We also transfer using \cite{attensionTransfer} in ``\cite{attensionTransfer}+vgg-Jigsaw++''. In ``CC+HOG'', we cluster bag of words of HOG and predict them with AlexNet; this is not fully unsupervised since HOG is hand-crafted. Surprisingly, it outperforms many SSL algorithms on classification.}
%We use $acc$ and $vcc$ to denote AlexNet (a) or VGG16 (v) pre-training and cluster classification (cc) on $2$K pseudo-labels from ImageNet. {\color{red} The last row shows the performance of the attention transfer method~\cite{attensionTransfer} where we solve the jigsaw++ task but also mimic the activation of the jigsaw++ task trained on VGG network.} }
%As in Table~\ref{tbl:voc} the second row uses a smaller stride on the first layer and the third row uses the groupless AlexNet.
%The methods are grouped based on the same scheme as in Table~\ref{tbl:voc}.
 } \label{tbl:voc}
\begin{adjustbox}{width=.48\textwidth}
\begin{tabular}{@{}l@{\hspace{.5em}}c c  c c   c }%c@{\hspace{.5em}}c@{}}
\toprule %\hline
\textbf{Method}  & \textbf{Ref} & \textbf{Class.} & \multicolumn{2}{c}{\textbf{Det.}} & \textbf{Segm.} \\
& & & \textbf{SS} & \textbf{MS} & \\
\midrule%\hline
Supervised~\cite{AlexNet12}  &  & 79.9 & 59.1 & 59.8 & 48.0 \\% &
\hline
CC+HOG~\cite{hog} &   & 70.2  & 53.2 & 53.5 & 39.2 \\
\hline
Random  	&  \cite{ContextEncoder} & 53.3 & 43.4 & - & 19.8 \\
%ego-motion~\cite{agrawal15}  	& \cite{agrawal15}  & 52.9 & 41.8 &  \\ 
ego-motion~\cite{agrawal15}   	& \cite{agrawal15}  & 54.2 &  43.9 &-  & - \\
BiGAN~\cite{advLearning} & \cite{advLearning}  & 58.6 & 46.2 & - & 34.9 \\
ContextEncoder~\cite{ContextEncoder} & \cite{ContextEncoder}  & 56.5 & 44.5 & -  & 29.7  \\
%Sound~\cite{ambientSound}   &  \cite{splitBrain} & 54.4 & 44.0 & - \\
%Sound~\cite{ambientSound}   &  \cite{splitBrain} & 61.3 & - & - \\
%Video~\cite{wangVideo}   &  \cite{krahenbuhl2015data}  &62.8 & 47.4  &  -\\
Video~\cite{wangVideo}  & \cite{krahenbuhl2015data}  & 63.1 &  47.2 & - & - \\
Colorization~\cite{colorful}  &  \cite{colorful} & 65.9  & 46.9 & - & 35.6 \\
Split-Brain~\cite{splitBrain}     & \cite{splitBrain} & 67.1 & 46.7 & - & 36.0  \\
Context~\cite{context}  	& \cite{krahenbuhl2015data}  & 55.3 & 46.6 & - & - \\
Context~\cite{context}$^*$  	& \cite{krahenbuhl2015data}  & 65.3 & 51.1 & - & -  \\
Counting~\cite{counting}  &  \cite{counting} & 67.7 & 51.4  & - & 36.6 \\
WatchingObjectsMove~\cite{WatchingObjects} &\cite{WatchingObjects} &  61.0 & - & 52.2 & -\\
Jigsaw~\cite{noroozi2016}    & \cite{noroozi2016} & 67.7 & 53.2 & - & - \\
Jigsaw++    &  & 69.8  & 55.5 & 55.7 & 38.1 \\
CC+Context-ColorDrop~\cite{context}  	&   & 67.9 &  52.8 & 53.4 & - \\
CC+Context-ColorProjection~\cite{context}  	&   & 66.7 & 51.5 & 51.8 & - \\
CC+Jigsaw++    &  & 69.9  & 55.0 &  55.8 & 40.0 \\
\cite{attensionTransfer}+vgg-Jigsaw++    &  & 70.6 & 54.8 & 55.2 & 38.0\\
CC+vgg-Context~\cite{context}  	&   & 68.0  & 53.0 & 53.5 & - \\
CC+vgg-Jigsaw++    &  & \textbf{72.5} & \textbf{56.5} & \textbf{57.2} & \textbf{42.6}\\
\bottomrule%\hline
\end{tabular}
\end{adjustbox}
\end{table}

\subsection{Ablation Studies}

%Fix pre-trained network (e.g., Jigsaw++ with occlusions Alex in ImageNet) and transfer to PASCAL VOC07

%All ablation studies on PASCAL VOC2007 classification.
%Pretraining on ImageNet with AlexNet in all stages.
%-> best setting for 2K
All the ablation studies are carried out with AlexNet pre-trained on the Jigsaw++ task with ImageNet as dataset (trainset without the labels). 
%Then, clustering is carried out on \texttt{conv4} features extracted from ImageNet images unless specified otherwise.
The pseudo-labels are also assigned to ImageNet data unless specified otherwise. The knowledge transfer is then completed by training AlexNet on the pseudo-labels. Finally, this model is fine-tuned on PASCAL VOC 2007 for object classification.\\
%Change the number of clusters in ImageNet, k = (500 done, 1K done, 2K done, 5K done, 10K done) train cluster-classifier on imagenet
\noindent\textbf{What is the impact of the number of clusters? }
The k-means clustering algorithm needs the user to choose the number of clusters. 
In principle, too few clusters will not lead to discriminative features and too many clusters will not generalize. Thus, we explore different choices to measure the sensitivity of our knowledge transfer algorithm. Since each cluster corresponds to a pseudo-label, we can loosely say that the number of clusters determines the number of object categories that the final network will be able to discern. Therefore, one might wonder if a network trained with very few pseudo-labels develops a worse learning than a network with a very large number of pseudo-labels. This analysis is analogous to work done on the ImageNet labels \cite{WhatMakedImageNetGood}. Indeed, as shown in Table~\ref{tab:clusters}, we find that the network is not too sensitive to the number of clusters. Perhaps, one aspect to further investigate is that, as the number of clusters increases, the number of data samples per cluster decreases. This decrease might cause the network to overfit and thus reduce its performance despite its finer categorization capabilities.\\
% Change clustering dataset: cluster on Places then cluster-classify on imagenet (all 3 left to do)
%\begin{table}[t]
%\caption{Impact of the cluster and pseudo-labels data domains}
%\label{tab:domain}
%\footnotesize
%\centering
%%\begin{adjustbox}{width=.48\textwidth}
%\begin{tabular}{@{} l @{\hspace{2.8em}}  c @{\hspace{1.5em}}  c @{}}
%\toprule%\hline
%\textbf{cluster/pseudo-labels}   & \textbf{ImageNet/ImageNet} & \textbf{Places/ImageNet} \\
%\textbf{voc-classification} & - & - \\
%\bottomrule%\hline
%\end{tabular}
%\end{table}
\noindent\textbf{What is the impact of the cluster data domain? }
Our knowledge transfer method is quite flexible. It allows us to pre-train on a dataset, cluster on another, and then define pseudo-labels on a third one. In this study, we explore some of these options to illustrate the different biases of the datasets. The results are shown in Table~\ref{tab:domain}. In all these experiments, we pre-train AlexNet on the Jigsaw++ task with ImageNet. Then, we decouple the training and inference of the clustering algorithm. For instance, in the right column of Table~\ref{tab:domain}, we learn cluster centers on \texttt{conv4} features of Jigsaw++ extracted on Places and then run the assignment of clustering on ImageNet to get pseudo-labels to be used in the training of the final AlexNet. We see only a small reduction in performance, which implies that our clustering method is not relying on particular biases inherent in the ImageNet dataset.\\

\begin{table}[t]
%\begin{center}
 \caption{{\textbf{ImageNet classification with a linear classifier.}} We use the publicly available code and configuration of \cite{colorful}. Every column shows the top-1 accuracy of AlexNet on the classification task. The learned weights from \texttt{conv1} up to the displayed layer are frozen. The features of each layer are spatially resized until there are fewer than 9K dimensions left. A fully connected layer followed by softmax is trained on a 1000-way object classification task.  
%As in Table~\ref{tbl:voc} the second row uses a smaller stride on the first layer and the third row uses the groupless AlexNet.
%The methods are grouped based on the same scheme as in Table~\ref{tbl:voc}.
 } \label{tbl:imagenet_lin}
\begin{adjustbox}{width=.48\textwidth}
\begin{tabular}{@{}l  c  c  c   c c c@{}}
\toprule%\hline
\textbf{Method}   & \textbf{Ref} & \textbf{conv1} & \textbf{conv2} & \textbf{conv3} & \textbf{conv4} & \textbf{conv5}\\
\midrule%\hline
Supervised~\cite{AlexNet12} & \cite{splitBrain} & 			19.3 & 36.3 & 44.2 & 48.3 & 50.5\\
\hline
CC+HOG~\cite{hog}  & & 						16.8  & 27.4 & 20.7 &  32.0 & 29.1 \\
\hline
Random  & \cite{splitBrain}  & 								11.6 & 17.1 & 16.9 & 16.3 & 14.1 \\
Context~\cite{context}    & \cite{splitBrain} & 			16.2 & 23.3 & 30.2 & 31.7 & 29.6\\ 
ContextEncoder~\cite{ContextEncoder} &\cite{splitBrain} & 	14.1 & 20.7 & 21.0 & 19.8 & 15.5\\
BiGAN~\cite{advLearning} &\cite{splitBrain} & 		17.7 & 24.5 & 31.0 & 29.9 & 28.0  \\
Colorization~\cite{colorful}   &  \cite{splitBrain} & 		12.5 & 24.5 & 30.4 & 31.5 & 30.3 \\
Split-Brain~\cite{splitBrain}     & \cite{splitBrain} & 	17.7 & 29.3 & 35.4 & 35.2 & {32.8} \\
%Counting(COCO)	\\
Counting~\cite{counting}& \cite{counting} & 								18.0 & 30.6 & 34.3 & 32.5 & 25.7\\
%Jigsaw~\cite{norooziArXiv2017} & \cite{norooziArXiv2017} & 	{18.2}& 28.8 & 34.0 & \underline{33.9} & 27.1 \\
Jigsaw++ & 			&			 18.2 & 28.7  & 34.1  & 33.2 & 28.0 \\
CC+Jigsaw++ & 	&					{18.9} & {30.5} & {35.7} & {35.4} & 32.2   \\
CC+vgg-Jigsaw++ & 		&				\textbf{19.2} & \textbf{32.0}  & \textbf{37.3} & \textbf{37.1} & \textbf{34.6} \\
\bottomrule%\hline
\end{tabular}
\end{adjustbox}
\end{table}

\subsection{Transfer Learning Evaluation}

We evaluate the features learned with different SSL methods on PASCAL VOC for object classification, detection, and semantic segmentation. Also, we apply our knowledge transfer method to some of these SSL methods under relevant settings. In particular, we apply our knowledge transfer to: Context \cite{context}, Context-ColorDropping \cite{context}, Context-ColorProjection \cite{context}, Jigsaw++, and HOG \cite{hog}.

%Fix pre-trained network (e.g., Jigsaw++ with occlusions Alex in ImageNet) and transfer to PASCAL VOC07

\begin{table}[t]
%\begin{center}
 \caption{{\textbf{Places classification with a linear classifier.}} {We use the same setting as in Table~\ref{tbl:imagenet_lin} except that to evaluate generalization across datasets, the model is pre-trained on ImageNet (with no labels) and then tested with frozen layers on Places (with labels). %The last layer has 205 neurons for scene categories.
}
%We use the publicly available code and configuration of \cite{colorful}. Every column shows the top-1 accuracy of AlexNet on the classification task. The learned weights from \texttt{conv1} up to the displayed layer are frozen. The features of each layer are spatially resized until there are fewer than 9K dimensions left. A fully connected layer followed by softmax is trained on a 205-way scene classification task.  
%As in Table 1 of the main paper, the third row uses a smaller stride on the first layer and the fourth row uses the group-less AlexNet. The numbers of all the other methods are taken from \cite{splitBrain}.
%The methods are grouped based on the same scheme as in Table~\ref{tbl:voc}.
} \label{tbl:places_lin}
\footnotesize
\centering
%\begin{adjustbox}{width=.48\textwidth}
\begin{tabular}{@{}l@{\hspace{2.8em}}  c@{\hspace{1.5em}}  c@{\hspace{1.5em}}   c@{\hspace{1.5em}} c@{\hspace{1.5em}} c@{}}
\toprule%\hline
\textbf{Method}   & \textbf{conv1} & \textbf{conv2} & \textbf{conv3} & \textbf{conv4} & \textbf{conv5}\\
\midrule%\hline
Places labels~\cite{places}  & 			22.1 & 35.1 & 40.2 & 43.3 & 44.6\\
ImageNet labels~\cite{AlexNet12}  & 	22.7 & 34.8 & 38.4 & 39.4 & 38.7\\
\hline
CC+HOG~\cite{hog} & 				20.3  & 30.0 & 31.8 &  32.5 & 29.8 \\
\hline
Random    &  							15.7 & 20.3 & 19.8 & 19.1 & 17.5 \\
Context~\cite{context}  & 				19.7 & 26.7 & 31.9 & 32.7 & 30.9\\ 
Jigsaw~\cite{norooziArXiv2017} &  {23.0} & 31.9 & 35.0 & 34.2 & 29.3 \\
Context encoder~\cite{ContextEncoder} & 18.2 & 23.2 & 23.4 & 21.9 & 18.4 \\
Sound~\cite{ambientSound}& 				19.9 & 29.3 & 32.1 & 28.8 & 29.8 \\
BiGAN~\cite{advLearning} & 		22.0 & 28.7 & 31.8 & 31.3 & 29.7  \\
Colorization~\cite{colorful}    & 		16.0 & 25.7 & 29.6 & 30.3 & 29.7 \\
Split-Brain~\cite{splitBrain}    & 		21.3 & 30.7 & 34.0 & 34.1 & 32.5 \\
%Counting(COCO)	\\
Counting~\cite{counting} & 				\textbf{23.3} & 33.9 & 36.3 & 34.7 & 29.6 \\
Jigsaw++ & 						 		22.0 & 31.2  & 34.3  & 33.9 & 22.9 \\
CC+Jigsaw++ & 						22.5 & 33.0 & 36.2 & {36.1} & {34.0}  \\
CC+vgg-Jigsaw++ & 						22.9 & \textbf{34.2}  & \textbf{37.5} & \textbf{37.1} &\textbf{34.4} \\
\bottomrule%\hline
\end{tabular}
%\end{adjustbox}
\end{table}

\noindent\textbf{Fine-Tuning on PASCAL VOC. }
In this set of experiments, we use fine-tuning on PASCAL VOC as a common benchmark for all the SSL methods. The comparisons are based on object classification and detection on VOC2007 using the framework of~\cite{krahenbuhl2015data} and Fast-RCNN~\cite{girshickICCV15fastrcnn} respectively. We also report semantic segmentation result on VOC2012 dataset using the framework of ~\cite{fcn}.
We found that in most recent SSL papers, the settings of the detection task used for the evaluation of SSL methods are not the same as the ones used for supervised learning. More specifically, most SSL methods are using multi-scale fine-tuning for $150$K iterations, with the basic learning rate of $0.001$ and dividing the learning rate by $10$ every $50$K iterations. Moreover, some of the methods have been evaluated using the multi-scale test. We found that fine-tuning supervised weights with these settings achieves $59.1\%$ and $59.9\%$ with multi scale and single scale test respectively. We believe that it would be useful to use these as the baseline. We follow the same settings in all of our evaluations and report the results for both cases. We have locked the first layer in all cases including supervised weights as it is the default settings of Fast-RCNN.
%and we use the same settings as in prior comparisons in the literature \cite{context,krahenbuhl2015data,colorful,splitBrain,counting}. 
%We also report the performance obtained with AlexNet pre-trained in a supervised way on ImageNet labels. In particular, we point out that the detection performance has been obtained under the same settings used by the SSL methods (multi-scale search both in training and testing). When instead both training and testing is done with a single scale search, the performance is $57.1$ mAP, as previously reported.
In Table~\ref{tbl:voc}, we use ``CC+'' when our knowledge transfer method is used and ``vgg-'' when VGG16 is used in pre-training. All methods in Table~\ref{tbl:voc} use AlexNet for cluster prediction and fine-tuning.
%to specify what options have been used. We use $acc$ and $vcc$ to denote AlexNet (a) or VGG16 (v) pre-training and cluster classification (cc) on $2$K pseudo-labels from ImageNet.
In the case of HOG features \cite{hog} we only apply the cluster classification on pseudo-labels obtained from HOG on ImageNet. Surprisingly, these handcrafted features yield a very high performance in all three tasks.
%We also apply our knowledge transfer to the \text{context} SSL method. We find that our knowledge transfer method does not affect its performance adversely. Moreover, we also find that by pre-training context SSL method {\hamed true?} on the VGG16 network \cite{VGG16} instead of AlexNet, there is a significant boost in performance on both classification and detection.
Our knowledge transfer method does not have a significant impact on the performance when the source and destination architectures are the same.
However, when pre-training on VGG16, there is a significant boost of $2.6\%$ in classification, $1.6\%$ in detection, and $2.6\%$ in semantic segmentation. These results show state-of-the-art performance on all tasks. More importantly, the gap between SSL methods and supervised learning methods is further shrinking by a significant margin. We believe that our method allows to use larger scale datasets and deeper models in pre-training, while still using AlexNet in fine-tuning. %We expect that our method could give a further boost in performance by pre-training on deeper networks and by using larger datasets.

\noindent\textbf{Linear Classification. }
We also evaluate the SSL methods by using a linear classifier on the features extracted from AlexNet at different convolutional layers \cite{splitBrain}. We apply this on both ImageNet \cite{imagenet} and Places \cite{places} and evaluate the classification performance on the respective datasets. We illustrate the performance on ImageNet in Table~\ref{tbl:imagenet_lin} and on Places in Table~\ref{tbl:places_lin}. 
As can be observed, the performance of Jigsaw++ is comparable to prior state-of-the-art methods. Surprisingly, our knowledge transfer method seems to be beneficial to the transferred model CC+Jigsaw++. Consistently with other experiments, we also observe that pre-training with VGG16 in CC+vgg-Jigsaw++ gives a further substantial boost (an average of almost 2\% improvement). We also notice that HOG features do not demonstrate a performance in line with the performance observed on PASCAL VOC. A similar scenario is observed on the Places dataset. Notice that the performance obtained with CC+vgg-Jigsaw++ is quite close to the performance achieved with supervised pre-training on ImageNet labels.
 
%{\color{red}As introduced by Zhang et al. [43], we train a linear classifier on top of the frozen layers on ImageNet [38] and Places [45] datasets. The results of these experiments are shown in Tables 2 and 3. Our method achieves a performance comparable to the other state-of-the-art methods on the ImageNet dataset and shows a significant improvement on the Places dataset. Training and testing a method on the same dataset type, although with separate sets and no labels, may be affected by dataset bias. To have a better assessment of the generalization properties of all the competing methods, we suggest (as shown in Table 3) using the Ima- geNet dataset for training and the Places benchmark for test- ing (or vice versa). Our method archives state-of-the-art results with the conv1-conv4 layers on the Places dataset. Interestingly, the performance of our conv1 layer is even higher than the one obtained with supervised learning when trained either on ImageNet or Places labels.
%}

\begin{table}[t]
%\begin{center}
 \caption{\textbf{ImageNet classification with a nonlinear classifier} as in \cite{noroozi2016}. Every column shows top-1 accuracy of AlexNet on the classification task. The learned weights from \texttt{conv1} up to the displayed layer are frozen. The rest of the network is randomly initialized and retrained. 
Notice that the reported results of \cite{wangVideo} are based on the original paper.
All evaluations are done with $10$ croppings per image.
%The methods are grouped based on the same setting differentiation as in Table~\ref{tbl:voc}.
} \label{tbl:imagenet_non}
\begin{adjustbox}{width=.475\textwidth}
\begin{tabular}{@{}l  c  c  c   c c c c c@{}}
\toprule%\hline
\textbf{Method}  & \textbf{Ref} & \textbf{conv1} & \textbf{conv2} & \textbf{conv3} & \textbf{conv4} & \textbf{conv5} & \textbf{fc6} & \textbf{fc7} \\
\midrule%\hline
Supervised~\cite{AlexNet12} & \cite{AlexNet12} & 57.3 & 57.3 & 57.3 & 57.3 & 57.3 \\
\hline
Random  & \cite{noroozi2016}  & 48.5 & 41.0 & 34.8 & 27.1 & 12.0 & - & -   \\
Video~\cite{wangVideo} & \cite{noroozi2016} & 51.8 & 46.9 & 42.8 & 38.8 & 29.8 \\
BiGAN~\cite{advLearning} & \cite{advLearning} & 55.3 & {53.2} & {49.3} & {44.4} & {34.9}  & - & -   \\
%Colorization\cite{colorful}   &  \cite{colorful} & - & 46.6 & 43.5 & 40.7 & \textbf{35.2} \\
%Counting(COCO)	& & 53.1 & 47.9 & 41.2 & 34.7 & 25.9\\
Counting~\cite{counting} & \cite{counting}& 54.7 & 52.7 & 48.2 & 43.3 & 32.9 & - & -   \\
Context~\cite{context}    & \cite{noroozi2016} & 53.1 & 47.6 & 48.7 & 45.6 & 30.4 & - & -   \\ 
Jigsaw~\cite{noroozi2016} & \cite{noroozi2016} & 54.7 &  52.8 & 49.7 & 45.3 & 34.6 & - & -   \\
Jigsaw++ & & 54.7 & 52.9 & 50.3 & 46.1 & 35.4 & - & -   \\
CC+-vgg-Context & & 55.0 & 52.0 & 48.2 & 44.3 & 37.9 & 29.1 &  20.3 \\
CC+Jigsaw++ & & 55.3  &  52.2  &  51.4  &  47.6 &    41.1 &  33.9         & 25.9  \\
CC+vgg-Jigsaw++ & & \textbf{55.9} & \textbf{55.1} & \textbf{52.4} & \textbf{49.5} & \textbf{43.9} & \textbf{37.3} & \textbf{27.9}\\
\bottomrule%\hline
\end{tabular}
\end{adjustbox}
\end{table}

\begin{figure*}[t]
\begin{center}
%\fbox{\rule{0pt}{.4\linewidth}\rule{\linewidth}{0pt}}
	 \begin{minipage}{.32\textwidth}
        \centering
        \includegraphics[width=1\linewidth ]{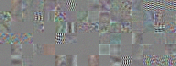} \\
        (a) %{\scriptsize{random}}
    \end{minipage} %
    \begin{minipage}{0.32\textwidth}
        \centering
        \includegraphics[width=1\linewidth]{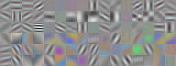} \\
        (b) %{\scriptsize{HOG}}
    \end{minipage}
    \begin{minipage}{0.32\textwidth}
        \centering
        \includegraphics[width=1\linewidth]{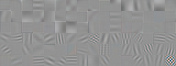} \\
		(c) %{\scriptsize{ContextColorDropping$^{acc}$}}
    \end{minipage}
    
    \begin{minipage}{0.32\textwidth}
        \centering
        \includegraphics[width=1\linewidth]{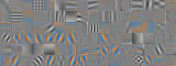} \\
        (d) %{\scriptsize{ContextColorProjection$^{acc}$}}
    \end{minipage}
        \begin{minipage}{0.32\textwidth}
        \centering
        \includegraphics[width=1\linewidth]{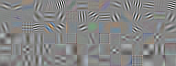} \\
        (e) %{\scriptsize{jigsaw++}}
    \end{minipage}
    \begin{minipage}{0.32\textwidth}
        \centering
        \includegraphics[width=1\linewidth]{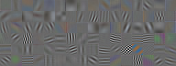} \\
        (f) %{\scriptsize{jigsaw++$^{acc}$}}
    \end{minipage}
    \vspace{-2mm}
\end{center}
   \caption{\texttt{conv1} filters of the cluster classification network using the AlexNet architecture trained with different pseudo-labels obtained from: (a) a randomly initialized AlexNet network, (b) CC+HOG, (c) Doersch \etal~\cite{context} trained with ColorDropping, (d)  Doersch \etal ~\cite{context} trained with ColorProjection, (e) CC+Jigsaw++ task trained on AlexNet, (f) the CC+vgg-Jigsaw++ task trained on VGG16.}
\label{fig:filters}
\end{figure*}

\begin{figure}[t]
\begin{center}
	\includegraphics[width=1\linewidth]{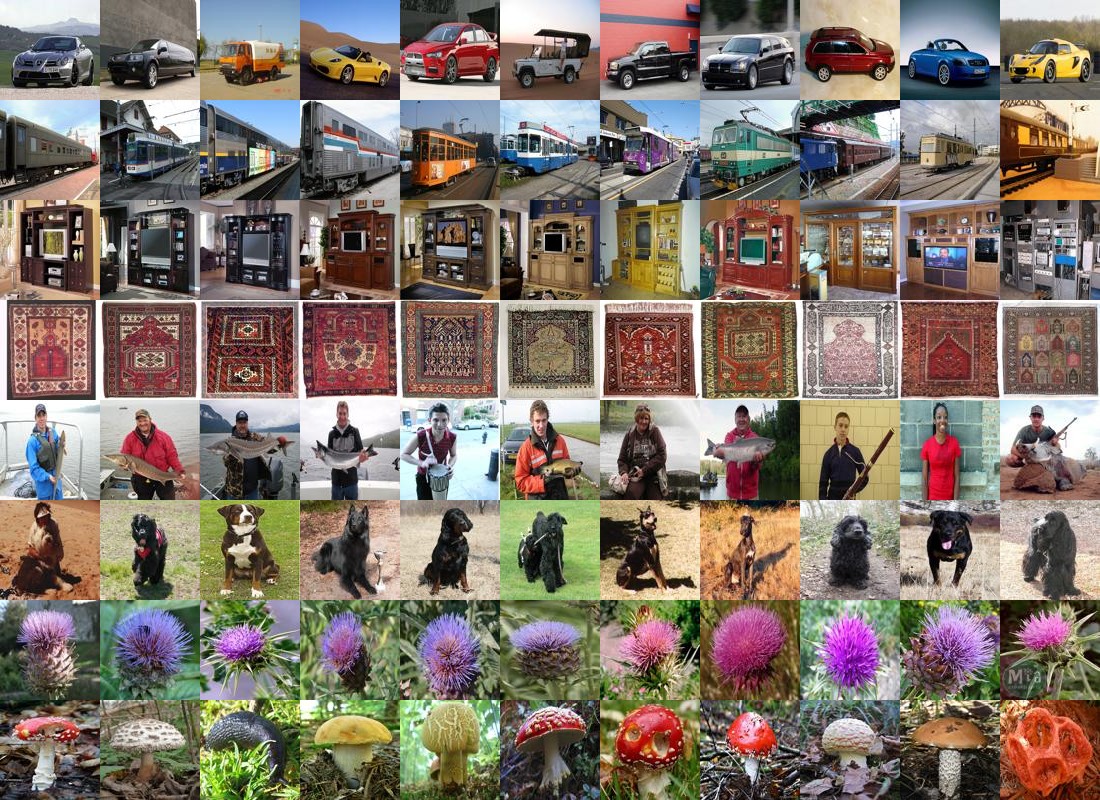}
\end{center}
   \caption{Some cluster samples used to train CC+vgg-Jigsaw++. Each row shows the $11$ closest images to their corresponding cluster center. }
\label{fig:vis}
\end{figure}

\noindent\textbf{Nonlinear Classification. }
We freeze several layers initialized with evaluating weights and retrain the remaining layers from scratch.
For completeness, we also evaluate SSL features as done in \cite{noroozi2016}, by freezing a few initial layers and training the remaining layers from random initialization. In comparison to the previous experiments, the main difference is that here we use a nonlinear classifier that consists of the final layers of AlexNet. This experiments is another way to illustrate the alignment between the pseudo-labels obtained from cluster classification and the ground truth labels. We show the comparisons in Table~\ref{tbl:imagenet_non}. The performance obtained by most SSL methods seems to confirm the pattern observed in the previous experiments. However, the difference between the previous state-of-the-art and CC+vgg-Jigsaw++ is quite remarkable. Also, the boost in performance of other prior work such as \cite{context} through pre-training with VGG16 is up to 9\% at \texttt{conv5}.

%{\color{red}Yosinski et al. [37] have shown that the last layers of AlexNet are specific to the task and dataset used for training, while the first layers are general-purpose. In the context of transfer learning, this transition from general-purpose to task- specific determines where in the network one should extract the features. In this section we try to understand where this transition occurs in the CFN. We repurpose the CFN, [9], and [36] to the classification task on the ImageNet 2012 dataset [8] and Table 2 summarizes the results on the validation set. The analysis consists of training each network with the labeled data from ImageNet 2012 by locking a subset of the layers and by initializing the unlocked layers with random values. If we train AlexNet we obtain the reference maximum accuracy of 57.4\%. Our method achieves 38.1\% when only fully connected layers are trained, which is 7.0\% higher than the next best performing algorithm [9]. There is a significant improvement (from 38.1\% to 48.3\%) when the conv5 layer is also trained. This shows that the conv5 layer starts to be specialized in the jigsaw puzzle reassembly task.}

\subsection{Visualizations}
We show some filters of the cluster classifier network in Figure~\ref{fig:filters}. We can see the impact of the pre-trained network on the cluster classifier. Interestingly, there is no color in the filters of the ``color dropping'' method of \cite{context} after knowledge transfer. This is consistent with the fact that this method does not see any color image in the pre-training stage. 
We also show some sample clusters used in training CC+vgg-Jigsaw++ in Figure~\ref{fig:vis}. Each row corresponds to images closest to the center of a single cluster. Ideally, for high-quality representations, we expect images from the same category on each row.

%
%\begin{figure}[t]
%\begin{center}
%	\setlength{\fboxsep}{0pt}
%	\setlength{\fboxrule}{1pt}
%	\hspace{-7cm}
%	\fcolorbox{red}{white}{
%	\begin{minipage}{.08\linewidth}
%	\includegraphics[height=1\linewidth,trim={2.5cm 1.6cm 29cm 1.1cm},clip]{figures/separate_rows/14.png}\\
%	\includegraphics[height=1\linewidth,trim={2.5cm 1.6cm 29cm 1.1cm},clip]{figures/separate_rows/13.png}
%	\end{minipage}}%
%	\fcolorbox{white}{white}{
%	\begin{minipage}{.08\linewidth}
%	\includegraphics[height=1\linewidth,trim={5.2cm 1.6cm 2.5cm 1.1cm},clip]{figures/separate_rows/14.png}\\
%    \includegraphics[height=1\linewidth,trim={5.2cm 1.6cm 2.5cm 1.1cm},clip ]{figures/separate_rows/13.png}
%	\end{minipage}}%
%\end{center}
%
%   \caption{\texttt{conv1} filters of cluster classification network using Alexnet architecture trained with different pseudo-networks obtained from: (a) randomly initialized AlexNet network, (b) HOG features, (c) Doersch \etal~\cite{context} trained with color dropping, (d)  Doersch \etal ~\cite{context} trained with color projection, (e) jigsaw++ task trained on AlexNet, (f) jigsaw++ task trained on VGG16.}
%\label{fig:vis}
%\end{figure}

\section{Conclusions}
Self-supervised learning is an attractive research area in computer vision since unlabeled data is abundantly available and supervised learning has serious issues with scaling to large datasets. Most recent SSL algorithms are restricted to using the same network architecture in both the pre-training task and the final fine-tuning task. This limits our ability in using large scale datasets due to limited capacity of the final task model. We have relaxed this constraint by decoupling the pre-training model and the final task model by developing a simple but efficient knowledge transfer method based on clustering the learned features. Moreover, to truly show the benefit, we increase the complexity of a known SSL algorithm, the jigsaw task, and use a VGG network to solve it. We show that after applying our ideas to transfer the knowledge back to AlexNet, it outperforms all state-of-the-art SSL models with a good margin shrinking the gap between supervised and SSL models from \%5.9 to \%2.6 on PASCAL VOC 2007 object detection task.

\noindent\textbf{Acknowledgements.} PF has been supported by the Swiss National Science Foundation (SNSF) grant number 200021\_169622. HP has been supported by GE Research and Verisk Analytics.

{\small
\bibliographystyle{ieee}
\bibliography{1024}

\begin{thebibliography}{10}\itemsep=-1pt

\bibitem{agrawal15}
P.~Agrawal, J.~Carreira, and J.~Malik.
\newblock Learning to see by moving.
\newblock In {\em ICCV}, 2015.

\bibitem{DistillationCaruana}
L.~J. Ba and R.~Caruana.
\newblock Do deep nets really need to be deep?
\newblock In {\em NIPS}, 2014.

\bibitem{buechlerCVPR17}
B.~Brattoli, U.~B{\"u}chler, A.~S. Wahl, M.~E. Schwab, and B.~Ommer.
\newblock Lstm self-supervision for detailed behavior analysis.
\newblock In {\em Proceedings of the IEEE Conference on Computer Vision and
  Pattern Recognition (CVPR)}, 2017.

\bibitem{caruana}
C.~Bucila, R.~Caruana, and A.~Niculescu-Mizil.
\newblock Model compression.
\newblock In {\em KDD}, 2006.

\bibitem{cimpoi2016deep}
M.~Cimpoi, S.~Maji, I.~Kokkinos, and A.~Vedaldi.
\newblock Deep filter banks for texture recognition, description, and
  segmentation.
\newblock {\em ICCV}, 2016.

\bibitem{hog}
N.~Dalal and B.~Triggs.
\newblock Histograms of oriented gradients for human detection.
\newblock In {\em CVPR}, 2005.

\bibitem{context}
C.~Doersch, A.~Gupta, and A.~A. Efros.
\newblock Unsupervised visual representation learning by context prediction.
\newblock In {\em ICCV}, 2015.

\bibitem{doersch2017multi}
C.~Doersch and A.~Zisserman.
\newblock Multi-task self-supervised visual learning.
\newblock {\em ICCV}, 2017.

\bibitem{advLearning}
J.~Donahue, P.~Kr{\"a}henb{\"u}hl, and T.~Darrell.
\newblock Adversarial feature learning.
\newblock In {\em ICLR}, 2017.

\bibitem{DosovitskiyExemplar}
A.~Dosovitskiy, P.~Fischer, J.~T. Springenberg, M.~Riedmiller, and T.~Brox.
\newblock Discriminative unsupervised feature learning with exemplar
  convolutional neural networks.
\newblock {\em PAMI}, 2014.

\bibitem{girshickICCV15fastrcnn}
R.~Girshick.
\newblock Fast r-cnn.
\newblock In {\em ICCV}, 2015.

\bibitem{HintonDistillation}
G.~Hinton, O.~Vinyals, and J.~Dean.
\newblock Distilling the knowledge in a neural network.
\newblock In {\em NIPS}, 2014.

\bibitem{WhatMakedImageNetGood}
M.~Huh, P.~Agrawal, and A.~A. Efros.
\newblock What makes imagenet good for transfer learning?
\newblock In {\em NIPS LSCVS Workshop}, 2016.

\bibitem{batch_norm}
S.~Ioffe and C.~Szegedy.
\newblock Batch normalization: Accelerating deep network training by reducing
  internal covariate shift.
\newblock In {\em ICML}, 2015.

\bibitem{tiedEgomotion}
D.~Jayaraman and K.~Grauman.
\newblock Learning image representations tied to ego-motion.
\newblock In {\em ICCV}, 2015.

\bibitem{krahenbuhl2015data}
P.~Kr{\"a}henb{\"u}hl, C.~Doersch, J.~Donahue, and T.~Darrell.
\newblock Data-dependent initializations of convolutional neural networks.
\newblock In {\em ICLR}, 2016.

\bibitem{AlexNet12}
A.~Krizhevsky, I.~Sutskever, and G.~E. Hinton.
\newblock Imagenet classification with deep convolutional neural networks.
\newblock In {\em NIPS}. 2012.

\bibitem{larsson2016learning}
G.~Larsson, M.~Maire, and G.~Shakhnarovich.
\newblock Learning representations for automatic colorization.
\newblock In {\em ECCV}, 2016.

\bibitem{fcn}
J.~Long, E.~Shelhamer, and T.~Darrell.
\newblock Fully convolutional networks for semantic segmentation.
\newblock In {\em CVPR}, 2015.

\bibitem{shuffle}
I.~Misra, C.~L. Zitnick, and M.~Hebert.
\newblock Shuffle and learn: Unsupervised learning using temporal order
  verification.
\newblock In {\em ECCV}, 2016.

\bibitem{noroozi2016}
M.~Noroozi and P.~Favaro.
\newblock Unsupervised learning of visual representations by solving jigsaw
  puzzles.
\newblock In {\em ECCV}, 2016.

\bibitem{norooziArXiv2017}
M.~Noroozi and P.~Favaro.
\newblock Unsupervised learning of visual representations by solving jigsaw
  puzzles.
\newblock {\em arXiv preprint arXiv:1603.09246}, 2016.

\bibitem{counting}
M.~Noroozi, H.~Pirsiavash, and P.~Favaro.
\newblock Represenation learning by learning to count.
\newblock In {\em ICCV}, 2017.

\bibitem{Okanohara07}
D.~Okanohara and J.~Tsuji.
\newblock A discriminative language model with pseudo-negative samples.
\newblock In {\em ACL}, 2007.

\bibitem{ambientSound}
A.~Owens, J.~Wu, J.~H.~M. annd William T.~Freeman, and A.~Torralba.
\newblock Ambient sound provides supervision for visual learning.
\newblock In {\em ECCV}, 2016.

\bibitem{WatchingObjects}
D.~Pathak, R.~Girshick, P.~Dollár, T.~Darrell, and B.~Hariharan.
\newblock Learning features by watching objects move.
\newblock {\em arXiv preprint arXiv:1612.06370}, 2016.

\bibitem{ContextEncoder}
D.~Pathak, P.~Krahenbuhl, J.~Donahue, T.~Darrell, and A.~A. Efros.
\newblock Context encoders: Feature learning by inpainting.
\newblock In {\em CVPR}, 2016.

\bibitem{imagenet}
O.~Russakovsky, J.~Deng, H.~Su, J.~Krause, S.~Satheesh, S.~Ma, Z.~Huang,
  A.~Karpathy, A.~Khosla, M.~Bernstein, A.~C. Berg, and L.~Fei-Fei.
\newblock Imagenet large scale visual recognition challenge.
\newblock {\em IJCV}, 2015.

\bibitem{SchmidhuberCuriousExplorer}
J.~Schmidhuber.
\newblock Formal theory of creativity, fun, and intrinsic motivation.
\newblock {\em PAMI}, 2014.

\bibitem{VGG}
K.~Simonyan and A.~Zisserman.
\newblock Very deep convolutional networks for large-scale image recognitoin.
\newblock In {\em ICLR}, 2015.

\bibitem{DAE06}
P.~Vincent, H.~Larochelle, Y.~Bengio, and P.-A. Manzagol.
\newblock Extracting and composing robust features with denoising autoencoders.
\newblock In {\em ICML}, 2006.

\bibitem{wangVideo}
X.~Wang and A.~Gupta.
\newblock Unsupervised learning of visual representations using videos.
\newblock In {\em ICCV}, 2015.

\bibitem{wangTransitive}
X.~Wang, K.~He, and A.~Gupta.
\newblock Transitive invariance for self-supervised visual representation
  learning.
\newblock In {\em ICCV}, 2017.

\bibitem{Wang-2016-4848}
Y.~Wang and M.~Hebert.
\newblock Learning to learn: Model regression networks for easy small sample
  learning.
\newblock In {\em ECCV}, 2016.

\bibitem{R2A2_hashlearning}
Y.-X. Wang, L.~Gui, and M.~Hebert.
\newblock Few-shot hash learning for image retrieval.
\newblock In {\em ICCVW}, 2017.

\bibitem{R2A1_NIPS2016_6408}
Y.-X. Wang and M.~Hebert.
\newblock Learning from small sample sets by combining unsupervised
  meta-training with cnns.
\newblock In {\em NIPS}, 2016.

\bibitem{attensionTransfer}
S.~Zagoruyko and N.~Komodakis.
\newblock Paying more attention to attention: Improving the performance of
  convolutional neural networks via attention transfer.
\newblock In {\em ICLR}, 2017.

\bibitem{generalization}
C.~Zhang, S.~Bengio, M.~Hardt, B.~Recht, and O.~Vinyals.
\newblock Understanding deep learning requires rethinking generalization.
\newblock In {\em ICLR}, 2017.

\bibitem{colorful}
R.~Zhang, P.~Isola, and A.~A. Efros.
\newblock Colorful image colorization.
\newblock In {\em ECCV}, 2016.

\bibitem{splitBrain}
R.~Zhang, P.~Isola, and A.~A. Efros.
\newblock Split-brain autoencoders: Unsupervised learning by cross-channel
  prediction.
\newblock In {\em CVPR}, 2016.

\bibitem{places}
B.~Zhou, A.~Lapedriza, J.~Xiao, A.~Torralba, and A.~Oliva.
\newblock Learning deep features for scene recognition using places database.
\newblock In {\em NIPS}, 2014.

\end{thebibliography}
}

\end{document}